\documentclass[10pt, conference, letterpaper]{IEEEtran}

\IEEEoverridecommandlockouts
\usepackage{xspace}
\usepackage{amsmath,amsfonts,amsthm}
\usepackage{textcomp}
\usepackage{mathtools}
\usepackage{url}
\usepackage{subcaption}
\usepackage{tikz}
\usepackage{comment}
\usepackage[ruled,vlined, linesnumbered]{algorithm2e}

\newcommand{\lr}[3]{\left #1 {#3} \right #2}

\newcommand{\sys}{\textsc{Tubo}\xspace}

\newenvironment{tightlist}{
\begin{list}{$\bullet$}{
    \setlength{\topsep}{.1em}
    \setlength{\partopsep}{0in}
    \setlength{\parskip}{0in}
    \setlength{\itemsep}{0in}
    \setlength{\parsep}{0in}
    \setlength{\leftmargin}{1em}
    \setlength{\rightmargin}{0in}
    \setlength{\itemindent}{0in}
}}
{\end{list}}

\newcommand{\eg}{\text{e.g.,}\ }
\newcommand{\ie}{\text{i.e.,}\ }

\newcommand{\tinyskip}{\vspace{3pt}}

\newcommand{\mypar}[1]{\tinyskip\noindent\textbf{#1.}\xspace}

\newcommand*\myc[1]{%
\scalebox{0.78}{\begin{tikzpicture}[baseline=-3pt]
  \node[draw,circle,inner sep=0.5pt, fill=black] {\textcolor{white}{\textsf{\textbf{#1}}}};
\end{tikzpicture}}}
\usepackage{cite}
\usepackage{amsmath,amssymb,amsfonts}
\usepackage{algorithmic}
\usepackage{graphicx}
\usepackage{textcomp}
\usepackage{xcolor}
\def\BibTeX{{\rm B\kern-.05em{\sc i\kern-.025em b}\kern-.08em
    T\kern-.1667em\lower.7ex\hbox{E}\kern-.125emX}}
\begin{document}

\setlength{\floatsep}{12pt}
\captionsetup[table]{aboveskip=17pt}
\setlength{\textfloatsep}{13pt}



\title{\sys: A Tailored ML Framework for Reliable Network Traffic Forecasting}

\author{\IEEEauthorblockN{Zhihang Yuan, Leyang Xue, Waleed Ahsan, Mahesh K. Marina}
\IEEEauthorblockA{\textit{The University of Edinburgh}}
}

\maketitle

\begin{abstract}

Traffic forecasting based network operation optimization and management offers enormous promise but also presents significant challenges from traffic forecasting perspective.
While deep learning models have proven to be relatively more effective than traditional statistical methods for time series forecasting, their reliability is not satisfactory due to their inability to effectively handle unique characteristics of network traffic.
In particular, the burst and complex traffic patterns makes the existing models less reliable, as each type of deep learning model has limited capability in capturing traffic patterns.
To address this issue, we introduce \sys, a novel machine learning framework custom designed for reliable network traffic forecasting.
\sys features two key components: burst processing for handling significant traffic fluctuations and model selection for adapting to varying traffic patterns using a pool of models. 
A standout feature of \sys is its ability to provide deterministic predictions along with quantified uncertainty, which serves as a cue for identifying the most reliable forecasts.
Evaluations on three real-world network demand matrix (DM) datasets (Abilene, GEANT, and CERNET) show that \sys significantly outperforms existing methods on forecasting accuracy (by 4$\times$), and also achieves up to 94\% accuracy in burst occurrence forecasting.
Furthermore, we also consider traffic demand forecasting based proactive traffic engineering (TE) as a downstream use case.
Our results show that compared to reactive approaches and proactive TE using the best existing DM forecasting methods, proactive TE powered by \sys improves aggregated throughput by 9$\times$ and 3$\times$, respectively.

\end{abstract}

\let\svthefootnote\thefootnote
\newcommand\freefootnote[1]{%
  \let\thefootnote\relax%
  \footnotetext{#1}%
  \let\thefootnote\svthefootnote%
}

\freefootnote{\hrule\vskip 3pt Short version of this paper is presented at ICDCS 2025.}


\section{Introduction}\label{introduction}

The Internet has been one of the most important infrastructures for the digital world.
As outlined in industry reports~\cite{ciscoreport}, nearly two-thirds of the global population relies on the Internet access, presenting both opportunities and challenges in network operations and management.
On one hand, effectively utilizing recorded traffic data offers opportunities to optimize network management. One of the most impactful ways to leverage this data is through network traffic forecasting. Accurately predicting future traffic within a network is critically important for various network operations and management tasks, including video streaming~\cite{bouten2014network}, traffic engineering (TE)~\cite{singh2022traffic}, capacity planning~\cite{le2016joint} and quality of service provisioning~\cite{wassie2023traffic}.

However, reliable network traffic forecasting is challenging, rooted in the difficulty associated with tracking and predicting the traffic pattern dynamics. These patterns are influenced by several factors, including seasonal changes, where traffic volumes fluctuate based on the time of day, day of the week, or even every minute, reflecting network usage and congestion. Link failures can cause routing changes, which in turn can alter traffic patterns between origin-destination (OD) pairs. For example, in video streaming, dynamic adaptation to network conditions is essential to maintain quality. Additionally, bursts in traffic, often related to sudden changes in user behavior, can introduce bias and variability into traffic data. These bursts are typically sparse but can significantly impact network performance and forecasting accuracy. Addressing these challenges requires sophisticated forecasting methods that can adapt to dynamic and unpredictable changes in traffic patterns so as to ensure robust and reliable network management.

Network traffic demand matrix (DM) forecasting~\cite{jiang2022internet} is a generalized view of network traffic forecasting outlined above as it involves traffic prediction/forecasting across all OD-pairs in the network.
It can be formulated as a (multi-dimensional) time series forecasting problem, given that the DM provides a snapshot of the traffic volume between different OD node pairs in the network at a given time.
In addition, DM forecasting incorporates the networking domain specific challenges as outlined above. 
However, existing time series forecasting approaches are not reliable as they do not effectively cope with the dynamic patterns and fluctuations.
RNN~\cite{DBLP:conf/iclr/LiYS018,schuster1997bidirectional,DBLP:journals/corr/FlunkertSG17} are effective at capturing short-term temporal dependencies, but they typically struggle to capture long-term dependencies due to vanishing gradients.
Iterated predictions from LSTMs~\cite{jiang2022internet,hochreiter1997long} can accumulate errors over time, leading to significant inaccuracies with longer sequences.
Transformers~\cite{DBLP:conf/iclr/0001YCL00L22, lim2021temporal} sometimes struggle to capture positional information, which is crucial for time series forecasting due to time dependency \cite{zeng2023transformers}. An improper choice of transformer-based models can fail to learn key patterns and increase training costs. Linear models~\cite{zeng2023transformers} fail to capture non-linearity, and their overly simplistic nature is inadequate for capturing the highly dynamic patterns present in traffic data.

Motivated by the above, we set out to answer two critical questions: 
\begin{enumerate}
    \item How can we effectively manage the network DM characteristics, i.e., bursts and varying patterns?
    \item How can we adapt models to accurately forecast upcoming traffic DM given its recent history?
\end{enumerate}

To answer these questions, we present \sys, a new tailored machine learning (ML) framework we introduce for network DM forecasting that is designed around two main principles: (1) adapting to huge traffic fluctuations with {\em burst processing}; and (2) {\em model selection} to cater to varying DM patterns. The burst processing module identifies and isolates outliers from the recently observed DM history (i.e., the input to DM forecasting) and redirects the outliers to a dedicated burst occurrence forecasting module. This produces a burst indicator for future timestamps, which prompts a downstream application of DM forecasting (e.g., TE) to rely on current DM measurement when a burst is anticipated. 
The non-outliers undergo the training process provided by the models with additional data preprocessing of normalization.
\sys leverages multiple state-of-the-art (SOTA) time-series forecasting models of different type (\eg RNN and transformer).
At inference time, \sys features a post-training uncertainty measurement and calibration capability to select the most suitable model estimated to yield the most accurate forecast based only on test input.

Our evaluation of \sys\ using three unique real-world network DM datasets (Abilene, GEANT, and CERNET) shows that \sys's approach to DM forecasting outperforms prior methods by 4$\times$, and achieves a high burst occurrence forecasting accuracy of up to 94\%. As a downstream use case to demonstrate the effectiveness of \sys, we consider TE. Our results show that forecasting based proactive TE powered by \sys\ significantly improves TE performance relative to the reactive approach as well as proactive TE with the best existing DM forecasting method by 9$\times$ and 3$\times$, respectively.

In summary, our contributions are as follows:
\begin{tightlist}
    \item~\emph{New and Reliable ML framework for DM forecasting (\S\ref{sec:framework})}: We address the complexities in network traffic, characterized by dynamic and bursty patterns, by proposing a novel DM forecasting framework called \sys. Crucially, \sys\ embeds careful burst handling and online model selection to overcome the limitations of prior DM forecasting methods and individual time-series forecasting models which struggle with bursts, diverse traffic patterns, and distribution drift. Upon publication, we intend to open source the \sys framework to benefit the research community. 
    \item~\emph{Comprehensive evaluation (\S\ref{sec:evaluation})}: We utilize three real-world network traffic demand matrix datasets (Abilene, GEANT, and CERNET) to extensively evaluate \sys. Our results show significant (4$\times$) boosts in DM forecasting accuracy and robustness with \sys compared to prior methods (e.g., \cite{jiang2022internet}).
    \item~\emph{TE Use Case (\S\ref{sec:evaluation-e2e})}: We consider TE as a key downstream application of DM forecasting and make the case for shifting from the traditional reactive and oblivious routing to a proactive TE paradigm, leveraging forecasted DM for better network traffic delivery. In particular, we demonstrate significant (almost an order-of-magnitude) efficiency gain with proactive TE powered by \sys\ compared to the commonly used reactive approach.
\end{tightlist}

\section{Background and Related Work}\label{sec:related-work}

Time series forecasting involves predicting future values based on previously observed values, aiming to understand and anticipate trends and patterns.
Numerous statistical methods for time-series forecasting exist, such as Gaussian Process Regression~\cite{roberts2013gaussian}, traditional ARIMA~\cite{box2015time} class of methods, and Prophet~\cite{de2011forecasting}. However, these methods suffer from various limitations. For example, ARIMA models require linear, stationary data, making them less suitable for complex traffic patterns. Gaussian Process Regression often assumes that the time series follows a specific distribution, and Prophet struggles with multi-dimensional time series, which is a common scenario in traffic demand matrix forecasting.

Generally speaking, deep learning models are more effective than traditional methods mentioned above for network traffic forecasting because they do not rely on strong prior distribution assumptions and can handle more complex traffic data patterns. Advanced deep learning based time series forecasting methods have emerged in the recent past.
Recurrent neural network (RNN) based prediction models, such as those in \cite{DBLP:conf/iscc/LiuWYSGT19,DBLP:conf/cnsm/LazarisP19}, have been employed with the ability to integrate with multi-commodity flow problems. RNNs~\cite{DBLP:conf/iclr/LiYS018,schuster1997bidirectional,DBLP:journals/corr/FlunkertSG17} and LSTMs~\cite{jiang2022internet,hochreiter1997long} are effective at capturing temporal dependencies. However, RNNs often struggle with long-term data due to vanishing gradients, an issue that LSTMs address but at the cost of requiring substantial data and computational resources. Additionally, iterated predictions from LSTMs can accumulate errors over time, leading to significant inaccuracies in longer sequences \cite{zeng2023transformers}.

Similarly, transformer-based models such as Temporal Fusion Transformer (TFT)~\cite{lim2021temporal} and Crossformer~\cite{DBLP:conf/iclr/0001YCL00L22} are believed to suffer from several issues: (1) the attention mechanism in transformers is position-invariant, requiring extra positional encoding to capture time dependency, whose effectiveness remains problematic~\cite{zeng2023transformers}; (2) transformer-based models typically have a large number of parameters due to the presence of multiple dense layers, leading to computational overhead, especially when traffic data patterns are not highly complex. Furthermore, linear models such as DLinear~\cite{zeng2023transformers} fail to capture the dynamic patterns arising from user behaviors in traffic data, and their overly simplistic nature is inadequate for capturing the highly dynamic patterns present in network traffic.
Additionally, more recent works \cite{DBLP:conf/iclr/0001YCL00L22, zeng2023transformers, wu2022timesnet, ansari2024chronos} claim superiority in their respective domains. However, different models are better suited to different patterns, even within the same time series, which exhibits varying patterns.

In summary, while these models offer various innovations, they typically do not fully capture the diverse and intricate nature of network traffic. Therefore, to effectively handle bursty and diverse traffic patterns, more advanced solutions are needed.
\section{Problem Statement}
\subsection{Demand Matrix (DM) Forecasting}\label{sec:problem_forecasting}
Demand matrix (DM) forecasting generalizes network traffic/demand forecasting by automatically considering all OD-pairs and providing predictions for all pairs simultaneously. We define the demand matrix (DM) $d_t$ at timestamp (epoch) $t$ as a matrix of size $N\times N$, where $N$ is the number of nodes in the network.
Each element in the matrix $d_{i,j}^t$ can have non-zero traffic demand, representing the demand between node $i$ and node $j$ in the network.
Generally, the demand matrix is not symmetric i.e., $d_{i,j}^t \not= d_{j,i}^t$, reflecting the asymmetric nature of network applications.

Consider a sequence of demand matrices over time, $D = [d^1, d^2, \dots, d^i]$, spanning $i$ epochs.
Epoch lengths can vary depending on the network scenario. For example, in the case of Abilene~\cite{DBLP:conf/sigmetrics/ZhangRDG03} the time granularity is 15min, while it is 5min with GEANT~\cite{DBLP:journals/ccr/UhligQLB06}.
The subset of the demand matrix series in a time window of length $w$ can be represented as $D^{s, w} := [d^{s-w+1},\dots,d^{s}]$. 
With a forecasting model $f: D^{s,w} \mapsto \mathbb{R}^{N \times N}$, its prediction output can be denoted by $f(D^{s,w}) := d^{s+1}$, representing the predicted demand matrix for the next time epoch.

\subsection{Characteristics of Demand Matrix Data}\label{sec:problem_characteristics}
We now delve into the distinct characteristics of network traffic data, including periodicity, trends, and bursts (as defined in~\S\ref{sec:problem_forecasting}), by analyzing real-world demand matrix datasets, such as those from Abilene~\cite{DBLP:conf/sigmetrics/ZhangRDG03} and GEANT~\cite{DBLP:journals/ccr/UhligQLB06}.

Specifically, when addressing the demand matrix forecasting problem, it is crucial to consider the concept of traffic bursts, characterized as significantly higher traffic volumes occurring over short periods. In this work, we define a burst as an outlier exceeding the 99th percentile (2.576 standard deviations) within a time window, a standard method widely adopted for outlier detection~\cite{grubbs1969procedures}. For a single Origin-Destination (OD) pair ($i$, $j$), its demand time series is defined as $D_{i,j}^{s,w} := [d_{i,j}^{s-w+1}, \dots, d_{i,j}^{s}]$. A given demand $d_{i,j}^t \in D_{i,j}^{s,w}$ is classified as a burst if it deviates significantly from the typical traffic pattern: $d_{i,j}^t - \mu(D_{i,j}^{s,w}) > 2.576 \times \sigma(D_{i,j}^{s,w})$, where $\mu$ and $\sigma$ represent the mean and standard deviation of the time series, respectively.

\begin{table}[t]
    \centering
    \small
    \begin{tabular}{|c|c|c|c|c|}
       \hline
       Dataset  & Epochs & Burst & Median (Mbps) & Max (Mbps)\\
       \hline
       Abilene  & 48,096 & 2.21\% & 9.19 & 119,319 \\
       GEANT  & 10,769 & 1.29\% & 0.527 & 19,088\\
       CERNET & 9,999  & 0.842\% & 9.14 & 9,858\\
       \hline
    \end{tabular}
    \caption{Real-world traffic burst characteristics.}
    \label{tab:burst-analysis}
\end{table}

Table~\ref{tab:burst-analysis} summarizes the statistics of bursts seen in three real-world DM datasets. We observe that the percentage of bursts is 1.29\% in GEANT. Despite their relatively low frequency, the magnitude of these bursts in terms of traffic demand is substantial. In certain instances, as observed in the Abilene network, there can be more than a $4.1 \times 10^5$ times increase in demand during a burst. \emph{Accurate accounting and handling of these bursts is therefore essential to guarantee the robustness of the forecasting algorithm, as these can otherwise lead to large traffic losses and performance degradation due to misprediction.}

\subsection{Challenges for Demand Matrix Forecasting}\label{sec:problem_challenges}

From the in-depth analysis of demand matrix characteristics, we have identified two primary challenges outlined below that significantly impact the accuracy and reliability of demand matrix forecasting. 
Addressing these challenges is crucial to achieve robust demand matrix forecasting for use in traffic engineering and beyond.

\mypar{Adaptively selecting prediction models based on time-varying and unseen demand patterns}
Many works focus on optimizing and designing neural network architectures to improve time series forecasting performance on specific datasets. However, little attention has been given to the adaptive selection of predictors based on the underlying patterns of the data. Our observations indicate that there is no universal prediction model capable of excelling across all different time series patterns. Even within a single time series, patterns can vary over time, and existing models often struggle to adapt to these variations. Therefore, it is crucial to develop an adaptive model selection strategy that dynamically chooses the most suitable predictor based on the currently encountered demand pattern. This approach ensures that the forecasting model remains robust and accurate, even when faced with time-varying and previously unseen demand patterns.

\mypar{Handling hard-to-predict and rare traffic bursts} 
Another crucial challenge is the uncertainty associated with traffic bursts in demand matrices. Most forecasting algorithms are inherently biased towards frequently occurring patterns, as these are more prevalent in the training data. However, this bias is problematic when dealing with burst events, which are infrequent but have a substantial impact on forecasting accuracy. In time series forecasting, the uncertainty and errors often stem from these burst values, which are challenging to predict due to their sporadic nature.
To effectively address this challenge, a method should be specifically designed to forecast burst events, despite their rarity. 
This strategy involves identifying and segregating burst values from regular patterns during the training phase.
By doing so, the model can focus on learning the more regular, non-bursty patterns, which constitute the majority of the data.
The burst events, once identified, can then be addressed separately through specialized burst-prediction mechanisms, enhancing the overall forecasting accuracy and robustness.

\section{\sys DM Forecasting Framework}\label{sec:framework}

In this section, we introduce \sys, our novel framework for network traffic demand forecasting. \sys optimizes forecasting performance through systematic input demand preprocessing and subsequent predicted demand postprocessing.

\subsection{Overview}
\begin{figure}[t]
    \centering
    \includegraphics[page=1, width=\linewidth]{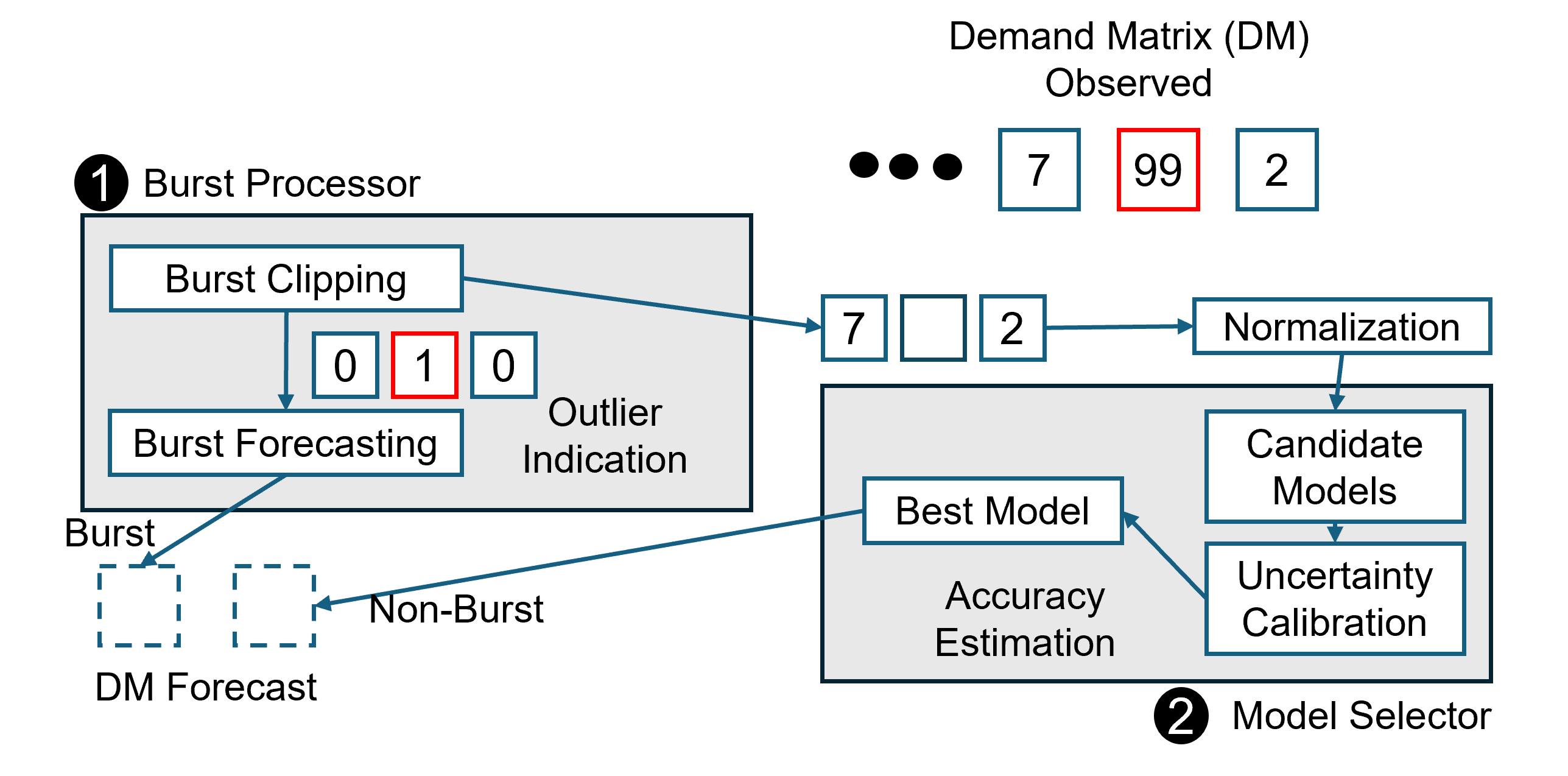}
    \caption{\sys\ framework illustrated.}
    \label{fig:overview}
\end{figure}

\sys\ is architected around two main components, each meticulously designed to address distinct aspects of traffic demand forecasting (see Fig.~\ref{fig:overview}) and corresponding to the two challenges highlighted earlier in \S\ref{sec:problem_challenges}: 

\mypar{Burst Processor} (\S\ref{sec:framework-burst}): This component is specifically tailored to handle bursts within the demand matrix data. It first segregates burst data from regular traffic data in the training dataset, to improve the predictability in the latter component. The burst and non-burst datasets are then separately used to train two distinct models for forecasting burst occurrences and DM values respectively, and their outputs are independently incorporated into the downstream applications. The burst processor (illustrated in \myc{1}) operates by first transforming the burst values into binary indicators, effectively denoting the presence or absence of a burst at each time point. For instance, given an input time series $[7,99,2]$, where `99' is an outlier, it is transformed into $[0,1,0]$. This binary time series is then utilized to predict forthcoming bursts. When a burst is anticipated, the processor skips the model selector component and simply forecasts a burst event for downstream applications. Assuming TE as an example downstream application, TE optimizer when informed of an impending burst event allocates sufficient bandwidth, ensuring the network is primed to handle sudden surges in data demand.

\mypar{Model Selector} (\S\ref{sec:framework-selection}): As per non-burst data,  we illustrate the model selector in \myc{2} to leverage the capability of multiple different forecasting models as an ensemble.
Formally, in our model selection process, we begin with a set of trained forecasting models, $\{f_1, f_2, \dots, f_m\}$, each capable of predicting the next demand matrix $d^{s+1}$ based on the input sequence $D^{s,w}$. For a given test input $D^{s,w}$, we utilize a selector function $g: \{f_1, f_2, \dots, f_m\} \mapsto f_k$ that evaluates each model $f_i$ and selects the model $f_k$ that is expected to provide the most reliable predictions for the specific pattern presented in the test input. This approach ensures that the selected model $f_k$ is best suited for the characteristics of the input sequence $D^{s,w}$, thereby optimizing the forecasting accuracy for the subsequent epoch.

Initially, the model selector trains a diverse pool of candidate models; each fine-tuned based on observed patterns and the chosen normalization technique. 
The input data is filtered to exclude bursts, ensuring models learn from the relatively predictable patterns.
For example, in the input time series $[7,99,2]$, the filtered data becomes $[0,1,0]$.
During the testing phase, upon the arrival of new test input (i.e., the most recent time window of DM time series), the model selector assesses the suitability of different models for the new input.
It uses the uncertainty level from each model to guide the selection process. By calculating the predictive uncertainty based on Monte Carlo Dropout \cite{gal2016dropout}, the model that exhibits the smallest calibrated uncertainty is chosen to forecast the traffic demand for the next epoch (time slot).
This approach not only guarantees precision but also enhances the framework's adaptability to varying data patterns.

In event of link failures and topology change, models typically require to have the capability of online learning, dynamic input size or to be retraining~\cite{liu2023leaf}.
The approach in \sys is orthogonal to the model capability since the updated model still requires data processing and post-training calibration and selection.


\subsection{Data Preprocessing}

The first step to the training phase in \sys\ is data preprocessing, as it segregates burst data from non-burst data and uses them to train separate models, as shown in Fig.~\ref{fig:overview}.
The detailed procedure is shown in Algorithm~\ref{algo:normalization}.

\mypar{Burst clipping} Burst values present two significant challenges to data processing and model training: (i) they can significantly distort the value scale of normalization, and (ii) they can lead to model bias, obscuring the underlying patterns essential for accurate forecasting. For instance, in z-score normalization, large bursts can inflate the mean and standard deviation. Furthermore, burst will make it more difficult for models to learn the most important patterns because burst will introduce more noisy values to the models during training. Common approaches in DM forecasting either indiscriminately incorporate burst values into the forecasting pipeline~\cite{DBLP:conf/infocom/GaoLCGGCG20} or arbitrarily clip them~\cite{jiang2022internet}.
\sys\ addresses the issue of burst values through both data preprocessing and model training. We observe that burst {\em values} are difficult to predict accurately with state-of-the-art time series forecasting models. Consequently, we detect bursts from the dataset and exclude them from value forecasting, reserving them only for \emph{occurrence} forecasting (Algorithm~\ref{algo:normalization} steps 2-9).
We use an empty value to fill in the space of burst data in the non-burst dataset such that during loss calculation this value can be ignored (step 6).

\begin{algorithm}[t]
    \KwIn{training dataset $D_T$, input window size $w$}
    \KwOut{dataset for burst $D_B$, and for non-burst $D_N$}

    \SetKwFunction{FData}{DataPreprocessing}
    \SetKwProg{Fn}{Function}{:}{}

    \Fn{\FData{$D_T, w$}}{
        \ForEach{$D^{s,w}$ in $D_T$}{
            $W_N := [], W_B := []$\;
            \ForEach{$d_{i,j}^t \in D^{s,w}$}{
                \If{$d_{i,j}^t - \mu(D^{s, w}_{i,j}) > 2.576 * \sigma(D^{s, w}_{i,j})$} {
                    $W_B$.append(1), $W_N$.append($N/A$)\;
                }
                \Else{
                    $W_B$.append(0), $W_N$.append($d_{i,j}^t$)\;
                }
            }
            $D_B$.append($W_B$), $D_N$.append($W_N$)\;
        }
        $D_B^{GLOB} := (D_N - \mu(D_N)) / \sigma(D_N)$\;
        $D_B^{INDV} := [(D_{i,j} - \mu(D_{i,j})) / \sigma(D_{i,j}), \forall i,j]$\;
        $D_B^{ROLL} := [(D_{i,j}^{s,w} - \mu(D_{i,j}^{s,w})) / \sigma(D_{i,j}^{s,w}), \forall i,j,s]$\;
        \Return $D_B$, $D_N = (D_B^{GLOB}, D_B^{INDV}, D_B^{ROLL})$\;   
    }
    \caption{Data Preprocessing Algorithm}
    \label{algo:normalization}
\end{algorithm}

\mypar{Normalization} Normalization is widely adopted as a data preprocessing step for model training as it ensures in our case uniformity across traffic demand scales for all OD pairs.
This uniformity is crucial for model training given its sensitivity to numeric scales and stability~\cite{DBLP:journals/gpem/Heaton18,10.1145/1015330.1015435}.
While normalization brings remarkable benefits to model training, there is currently a gap in its strategic application, particularly when it comes to traffic DM forecasting
In response to this need, \sys\ incorporates a normalization module with two primary design goals: (i) the capability to handle time series data with normalization in temporal order, maintaining the integrity of temporal patterns, and (ii) the ability to impute data with minimal change to the original time series data distribution, ensuring the validity of the model's learned patterns. 

We consider three types of normalization that present numeric stability and unique advantages.
At the same time, we make a key observation that any one of the methods does not outperform others on all datasets and models.
\begin{tightlist}
    \item \textit{GLOB}: The global approach normalizes every OD pair using the z-score derived from the entire training set. It treats the entire dataset as a single entity and adjusts the values of each OD pair relative to the mean and standard deviation of the whole set. (Algorithm~\ref{algo:normalization} step 10)
    \item \textit{INDV}: In the individual normalization approach, each OD pair is normalized using the z-score from its own specific subset within the training set. This allows each OD pair to be treated separately, adjusting the values relative to its own mean and standard deviation. (Algorithm~\ref{algo:normalization} step 11)
    \item \textit{ROLL}: In the rolling normalization approach, every OD pair is normalized using the z-score from each specific input. It continuously updates the mean and standard deviation based on the current window of input data, allowing the normalization process to adapt to recent changes in the data. (Algorithm~\ref{algo:normalization} step 12)
\end{tightlist}

By data preprocessing as described below, \sys\ ensures that data fed to the forecasting model is optimally prepared, laying the foundation for accurate and robust forecasting.

\begin{algorithm}[t]
\KwIn{DM dataset $D$, forecast input window size $w$, burst forecasting model $m_B$, set of models types $M_T$}

\SetKwFunction{FData}{DataPreprocessing}
\SetKwFunction{FTrain}{TrainModel}
\SetKwFunction{FKL}{KL-Divergence}
\SetKwFunction{FSelect}{ModelSelection}
\SetKwFunction{FSolver}{TESolver}
\SetKwProg{Fn}{Function}{:}{}

\For{each epoch $t$ in $D$}{
    Measure DM $d^t \in D$ for time slot $t$\;\label{step:te-measure}
    $D^{t,w} := [d^{t-w+1},\dots,d^{t}]$\;
    $D_B^t, D_N^t := $\FData{$D^{t,w}, w$}\;
    $b^{t+1} := m_B(D_B^t)$\;
    \uIf{$b^{t+1} == 1$}{
        Applications relies on real-time measurement\;
    }
    \Else{
        $m := $\FSelect{$D, M_T, w, D^{t,w}$}\;
        $d^{t+1} := m(D^{t,w})$\;
        Applications relies on forecasting output $d^{t+1}$\;
    }
    Add DM measurements to $D$\;
}
\caption{Forecasting pipeline with Burst}
\label{algo:burst-forecast-te}
\end{algorithm}

\subsection{Burst Occurence Forecasting}\label{sec:framework-burst}

As noted earlier, burst values in traffic demand data present two significant challenges to data processing and model training. We discussed our approach to data preprocessing to handle bursts above. Here we focus on burst occurence forecasting. 

In Algorithm~\ref{algo:burst-forecast-te}, we design the pipeline for seamlessly integrating burst forecasting into downstream applications.
This process keeps updating models online.
We align with the convention that defines bursts as outliers in a distribution (Algorithm~\ref{algo:normalization} step 5) and detect outliers for each window.
Following the detection, {\em DM time series of bursts is managed separately by binary forecasting}. After extracting the burst time series from the original DM, we transform the burst values into binary values, where $1$ indicates a burst, and $0$ indicates no burst (steps 6 and 8 in Algorithm~\ref{algo:normalization}). The burst forecasting algorithm only predicts whether a burst will occur in the next time slot (step~5). This binary forecasting approach is known for its accuracy compared to forecasting exact values (burst size in our setting)~\cite{DBLP:conf/cvpr/HeZRS16}.
When integrating the binary burst forecasting results with downstream applications, the algorithm indicates whether the applications should rely on forecasting output or real-time measurement (Algorithm~\ref{algo:burst-forecast-te} steps 6--11). This dual-path approach offers flexibility and ensures robustness against the unpredictable nature of burst.

\subsection{Model Selection}\label{sec:framework-selection}

The objective of our model selection design in \sys\ is to identify the model that yields the most reliable DM forecast. By ``reliable'', we mean maximizing accuracy while simultaneously minimizing uncertainty from two sources: (1) aleatoric uncertainty, which arises from inherent variability in the data, and (2) epistemic uncertainty, which stems from uncertainty in the model's predictions~\cite{amini2020deep}.
The underlying premise of our model selection strategy is that no universal model suits all scenarios, and the optimal choice often depends on the specific characteristics of the input. In this context, the level of uncertainty can serve as a valuable proxy variable for selecting the most appropriate model. However, it is crucial that the level of uncertainty, specifically the predictive $\sigma$, is correlated with our target metric, which in this case is the Mean Absolute Error (MAE).
In the design of \sys, forecast accuracy is not the sole criterion for model selection. The chosen model must also effectively handle the variations and complexities inherent in the DM data. Therefore, our model selection process is guided by both accuracy and the model's ability to adapt to changing patterns and its compatibility with the chosen normalization method, all of which are implicitly captured by the uncertainty quantification process. Furthermore, it is worth noting that the models in the model pool are subject to the users' choice, allowing them to select the models that best suit their specific scenarios.

\mypar{Quantifying uncertainty for each model}
There are two primary reasons for incorporating uncertainty quantification into the model selection: (i)~\emph{proximity to prediction accuracy}: different models excel at predicting certain time steps or patterns, and 
(ii)~\emph{decision resilience}: network operators can make more informed decision, ultimately enhancing network performance and reliability.


We achieve this by leveraging Monte Carlo (MC) Dropout~\cite{gal2016dropout}, a SOTA practical and scalable technique that approximates Bayesian inference in neural networks. 
MC Dropout enables the estimation of uncertainty by sampling from the model's predictive distribution.
This method balances computational efficiency and robust uncertainty quantification, making it suitable for traffic demand matrix forecasting, where the distribution of traffic demand is more complex and traditional methods fall short.
For a given forecasting model $f_i$, the predictive mean ($\mu_i$) and predictive uncertainty ($\sigma_i$) are estimated using MC Dropout. Specifically, we perform dropout during both the training and inference phases to generate multiple stochastic forward passes. This process allows us to sample from the model's predictive distribution.

Formally, let $D^{s,w}$ denote the input sequence, where $D^{s,w} = [d^{s-w+1}, \dots, d^s]$. The model $f_i$ produces a set of predictions $\{f_i^n(D^{s,w})\}_{n=1}^N$, where $N$ is the number of stochastic forward passes. The predictive mean ($\mu_i$) is calculated as:
$\mu_i = \frac{1}{N} \sum_{n=1}^N f_i^n(D^{s,w})$.
The predictive uncertainty ($\sigma_i$) is quantified as the standard deviation of these predictions:
$\sigma_i = \sqrt{\frac{1}{N} \sum_{n=1}^N (f_i^n(D^{s,w}) - \mu_i)^2}$.

\mypar{Improving the quality of uncertainty measure}
Accurate and comparable uncertainty quantification is critical for effective model selection.
It is widely observed that predictive distributions from regression models can often be miscalibrated~\cite{song2019distribution}, meaning that model can be overly certain about predictions with large error. This miscalibration can arise from various factors, such as the mismatch between the prior distribution of the loss function and the actual data distribution.
To address this issue, distribution calibration~\cite{song2019distribution} is employed to align uncertainty by enhance the correlation between the MAE and the predictive uncertainty ($\sigma$).
In addition, the value range of $\sigma$ from different models can also be comparable.

The distribution calibration aims to adjust the predicted uncertainty such that it more accurately reflects the true error observed in the forecasts.
This process involves recalibrating the predictive mean ($\mu$) and predictive uncertainty ($\sigma$) obtained from Monte Carlo Dropout as defined above. Formally, to achieve distribution calibration, we adopted the practice from \cite{song2019distribution}. Specifically, we utilize the stochastic variational inference \cite{hensman2015scalable} to transform the distributions output by the regressor.
Formally, the calibration function $h(\cdot)$ is learned on the training dataset together with the training process of the predictors from candidate model pool.
Let the calibrated predictive mean and uncertainty at the $i_{th}$ time step be denoted as $\hat{\mu}$ and $\hat{\sigma}$, respectively. The calibration process uses the original predictions $\mu$ and $\sigma$, along with the actual observed values from the test data. Through a calibration function, typically a form of regression, we adjust the predicted values to better match the observed outcomes. The calibration process can be expressed as:
$\hat{\mu_i}, \hat{\sigma_i} = h(\mu_i, \sigma_i, y_{\text{true}_i})$.
Here, the calibration function $h(\cdot)$ adjusts both the predictive mean and uncertainty based on the observed true values $y_{\text{true}}$. By improving this correlation, we ensure that the predictive uncertainty $\hat{\sigma_i}$ becomes a more reliable indicator of the model's forecasting error, thereby enhancing the overall robustness and reliability of the selected models.

\mypar{Selecting Models Based on Calibrated Uncertainty}
Since the calibrated model provides a stronger positive correlation to uncertainty and error of the prediction, we can thus select model with the smallest calibrated uncertainty.
These models are likely to perform best for the specific pattern presented in the test input, rather than solely aiming for high accuracy on the training and validation dataset conventionally.

The approach of selection is shown in Equation~\eqref{eq:model-selection}.
Given the all models with different type (\eg Crossformer, LSTM) and its corresponding training with three types of normalization, we aim to find the smallest $\sigma$ on test input $I$.
\begin{align}
    m_{\text{best}} := \min\limits_{\sigma}\lr\{\}{ f_{\text{calibrate}}(f_{\text{MC}}(m,I), I) }, \forall m\in M_T \label{eq:model-selection}
\end{align}
To quantify model uncertainty and improve the correlation between Mean Absolute Error (MAE) and predictive uncertainty (reflected by predictive $\sigma$, or $\hat{\sigma}$), we utilize the calibrated $\hat{\sigma}$ obtained from Monte Carlo Dropout~\cite{gal2016dropout}, \ie $f_{\text{MC}}$. 
The $f_{\text{MC}}$ provides a range of values that quantifies the uncertainty, the larger the variance is the higher the uncertainty.
The calibration function $f_{\text{calibrate}}$ can be trained offline suing validation dataset, and provide the calibrated $\sigma$ as output.
Since the inference speed only takes milliseconds and the calibrator is trained beforehand, it can satisfy real-time prediction requirements. Generally, we do not need an extensive number of models in the pool, as significant performance improvements can be achieved with as few as four models.

\mypar{Practical Concerns}
Regarding real-world deployment, there are two main practical concerns: the scalability of the \sys framework for large networks and the handling of demand matrix distribution shifts (a.k.a. concept drifts).

\noindent Firstly, the computational efficiency of TUBO is critical for large-scale network traffic forecasting. Using an NVIDIA H100 GPU with a peak performance of 500 TFLOPS, we evaluated the GPU computation cost for an input sequence of length 128, with a demand matrix of \(N\) nodes. The model operates at approximately 10 GFLOPS per node. Given TUBO's capacity to generate results for up to 20,000 nodes, predictions can be produced within 1-minute granularity. These findings demonstrate TUBO's scalability, enabling real-time forecasting and optimization for large networks. Secondly, regarding data distribution shifts, the most common approach is to retrain models tailored to the current data distribution \cite{zhang2024caravan}, though alternative methods, such as those proposed in \cite{liu2023leaf}, have also been explored. Resolving distribution shifts is a well-studied challenge with inherent complexities specific to the problem setting. We assume that the \sys framework, being a tailored demand matrix forecasting solution, addresses this as an orthogonal problem.

Additionally, one of the benefits of \sys's uncertainty quantification is its ability to handle unseen distribution patterns. As unseen data inherently carries more uncertainty, the use of Monte Carlo Dropout \cite{gal2016dropout} allows \sys to capture and reflect this uncertainty, thereby improving its robustness when encountering novel traffic patterns. This also alerts users to the potential distribution shift.

\section{Evaluation}\label{sec:evaluation}

In this section, we comprehensively evaluate \sys\ using real-world topologies and measurement datasets, focusing 
 on the traffic demand matrix. 
 We also carry out an extensive ablation study of our framework regarding network characteristics. 
Finally, we consider TE as a downstream use case and assess the TE performance gain resulting from enhanced forecasting accuracy with \sys. 

\subsection{Methodology}

\mypar{Datasets}
We evaluate the above methods using three well-established and publicly available DM datasets each representing different network topologies and characteristics. Abilene~\cite{DBLP:conf/sigmetrics/ZhangRDG03} dataset comprises of 12 nodes and 30 links, GEANT~\cite{DBLP:journals/ccr/UhligQLB06} dataset consists of 23 nodes and 74 links, and CERNET~\cite{jiang2022internet} dataset includes 14 nodes and 32 links. 
Each dataset provides traffic demand data for each OD pair at every epoch (timestamp), which is the basic input to the models without introducing new features to the datasets.
The timestamp granularity of Abilene and CERNET is 5 minutes whereas GEANT has a granularity of 15 minutes.
Each of the datasets is partitioned into two subsets: a training set (60\%) and test set (40\%).
Such data partition strategy is common for evaluating machine learning techniques~\cite{goodfellow2016deep}. All three datasets contain small amounts of missing data (62\% max for OD pair, 5-28\% on average) and burst (1-3\%). 
The calculation of error excludes the missing values in the test set, while training set includes missing values after zero imputation.

\mypar{Baseline Methods}
We evaluate a comprehensive spectrum of state-of-the-art (SOTA) time-series prediction methods in terms of their efficacy for DM forecasting. Specifically, we consider four different methods, spanning deep learning-based approaches (such as LSTM~\cite{DBLP:journals/neco/HochreiterS97}, 
GRU~\cite{DBLP:conf/emnlp/ChoMGBBSB14},
ConvLSTM~\cite{jiang2022internet}, Crossformer~\cite{DBLP:conf/iclr/0001YCL00L22}).
As a metric, we use mean absolute error (MAE), defined as the average absolute demand forecasting error across all OD pairs for the upcoming epoch, over all epochs.
We use all models in the model selection procedure.

\mypar{Training hyperparameters}
For each model, we train for a maximum of $100$ training epochs over the training dataset time epochs.
We use a K-fold validation on training set to choose the best model for each training, $k$ is chosen as the default value 10.
The optimization of models is executed using the Adam optimizer~\cite{adam}, with a learning rate of 0.002, betas set to (0.9, 0.98), and a weight decay of $10^{-5}$, ensuring a robust and efficient training process. 

\subsection{Accuracy of Demand Matrix Forecasting}\label{sec:evaluation-predict}

\begin{figure*}[t]
    \centering
    \begin{subfigure}{.32\linewidth}
        \includegraphics[width=\linewidth]{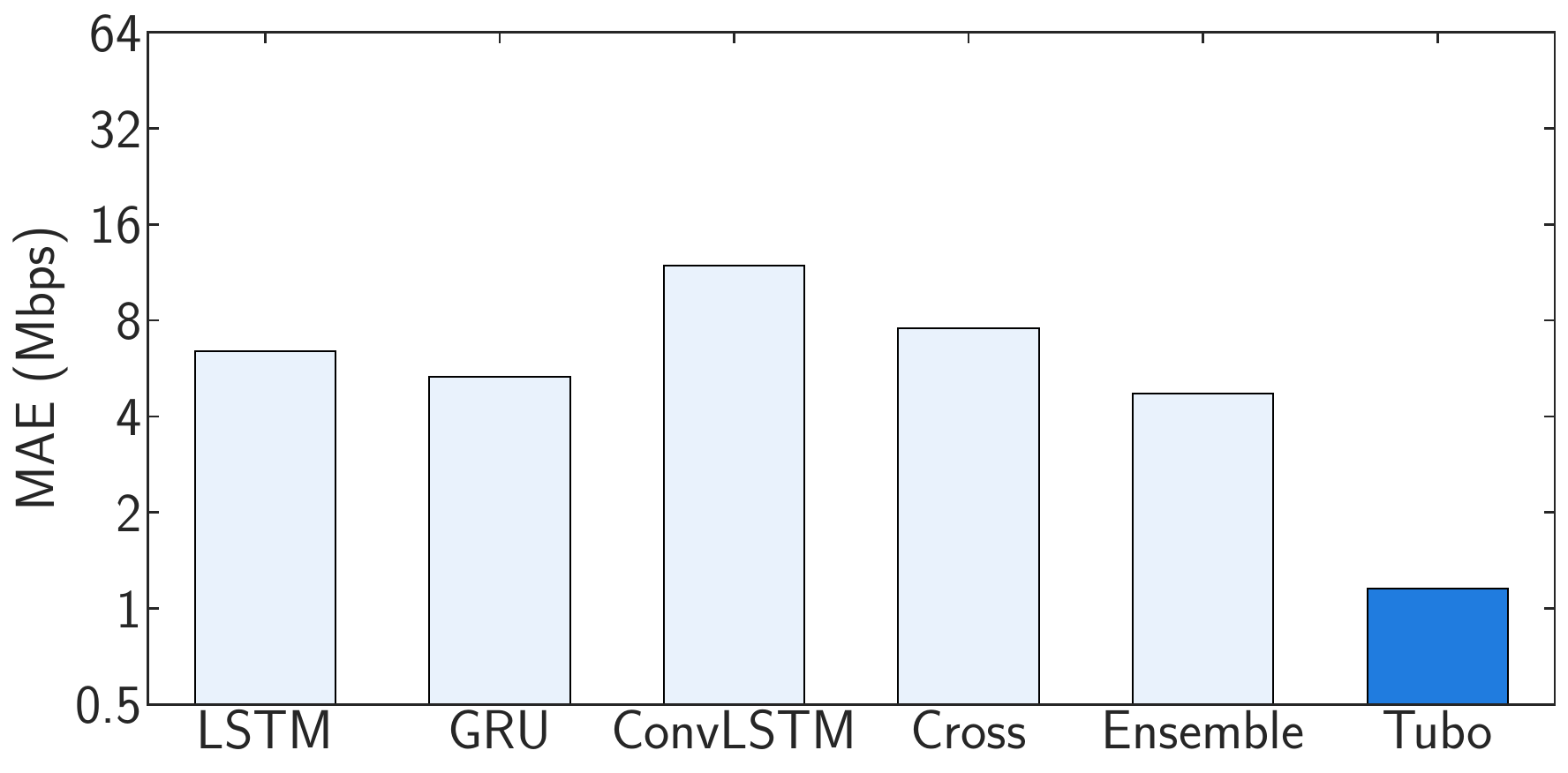}
        \vspace{-0.2in}
        \caption{Abilene}
        \label{fig:overall-mae-abilene}
    \end{subfigure}
    \begin{subfigure}{.32\linewidth}
        \includegraphics[width=\linewidth]{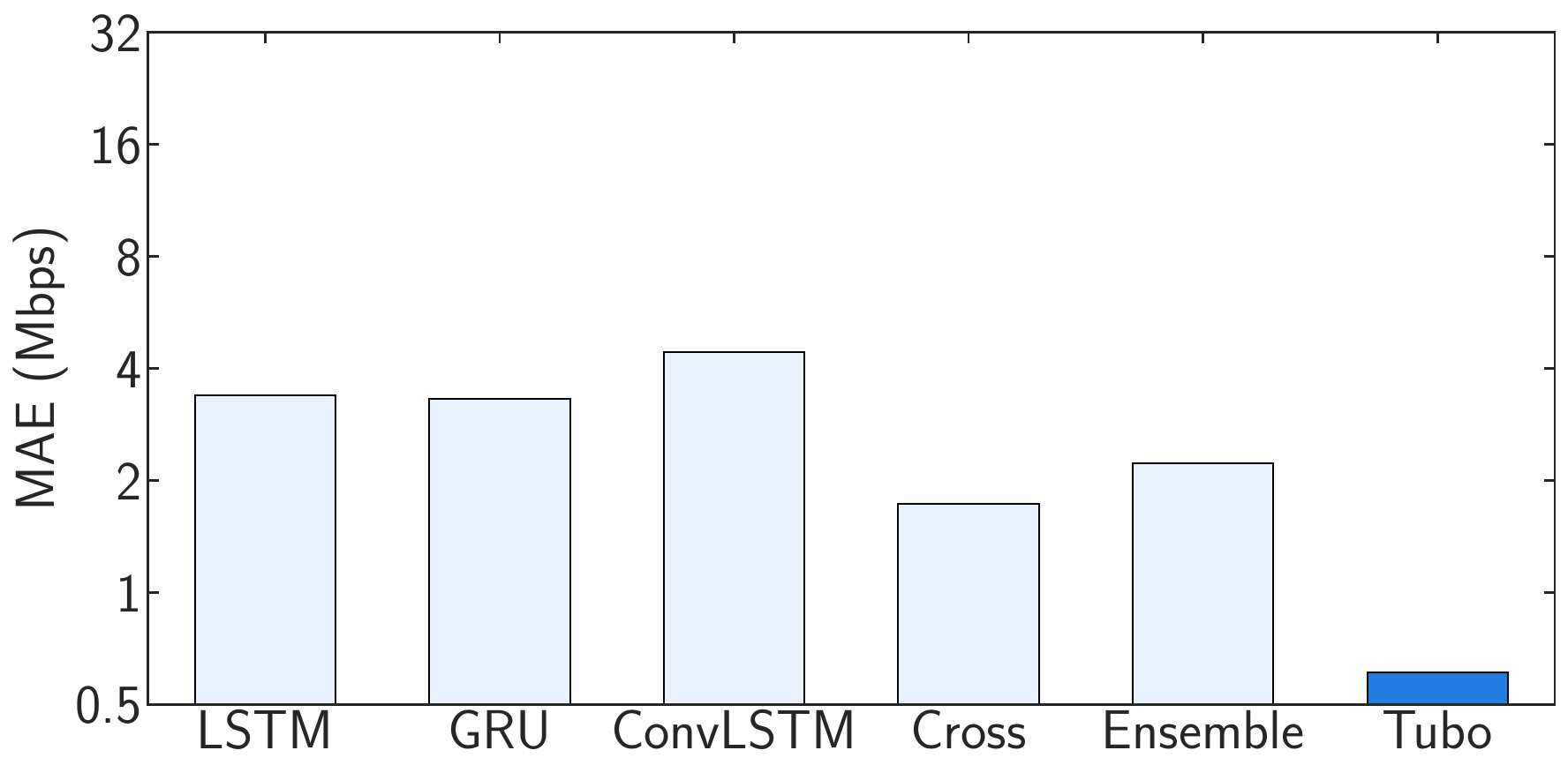}
        \vspace{-0.2in}
        \caption{GEANT}
        \label{fig:overall-mae-geant}
    \end{subfigure}
    \begin{subfigure}{.32\linewidth}
        \includegraphics[width=\linewidth]{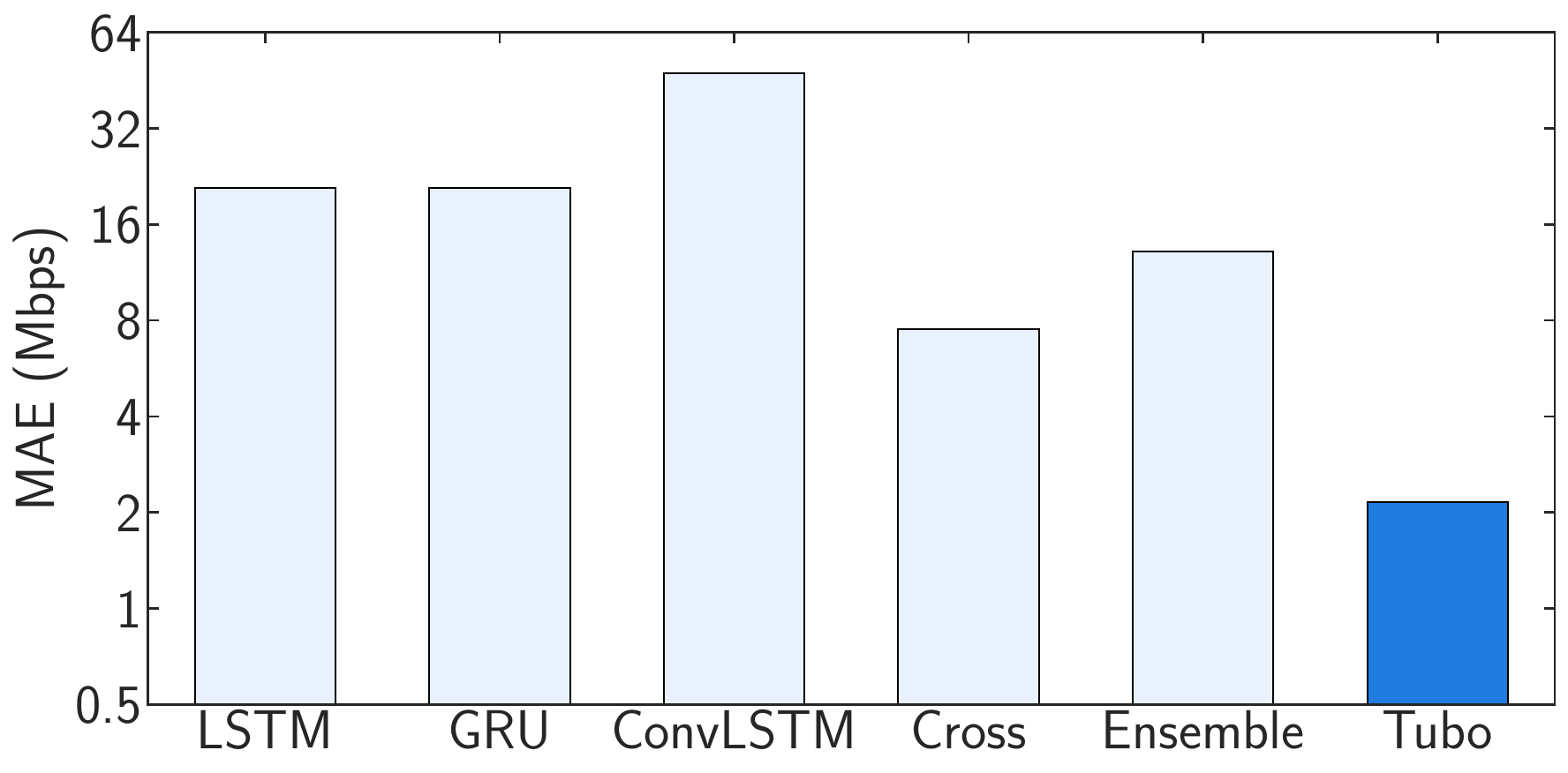}
        \vspace{-0.2in}
        \caption{CERNET}
        \label{fig:overall-mae-cernet}
    \end{subfigure}
    \vspace{-0.1in}
    \caption{MAE comparison among DM forecasting methods.}
    \label{fig:overall-mae}
\end{figure*}

\mypar{Mean Absolute Error (MAE)} When comparing the Mean Absolute Error (MAE) of various forecasting methodologies, as shown in Figures~\ref{fig:overall-mae-abilene},~\ref{fig:overall-mae-geant}, and~\ref{fig:overall-mae-cernet}, it becomes evident that \sys significantly surpasses baseline methodologies.
In all datasets evaluated, \sys outperforms deep learning (DL) based methods by 4$\times$, reducing MAE from 5.6 Mbps down to 1.4 Mbps.
The advancement of \sys can be partly attributed to the burst processor, the burst in the Abilene dataset is very significant (as shown in Table~\ref{tab:burst-analysis}), while the SOTA algorithm has been fitted to these large values with bias.
\sys can clip outbursts and tune to a more accurate model. 
The enhanced performance of \sys can be additionally attributed to the selection of appropriate forecasting models.
\sys integrates the capability of identifying interflow-correlation from Crossformer and the capability of identifying temporal locality from ConvLSTM. It is worth noting that even compared to the ensemble method, where the average of four models' predictions is used as the final predicted value, \sys\ still outperforms it. This superior performance can be attributed to the fact that simply averaging the models' outputs without considering the certainty level of their predictions is ineffective. Figures~\ref{fig:overall-mae-geant} and~\ref{fig:overall-mae-cernet} further evidence this, showing that the ensemble results are worse than those of single model performance.

\begin{figure*}[t]
    \centering
    \begin{subfigure}{.32\linewidth}
        \includegraphics[width=\linewidth]{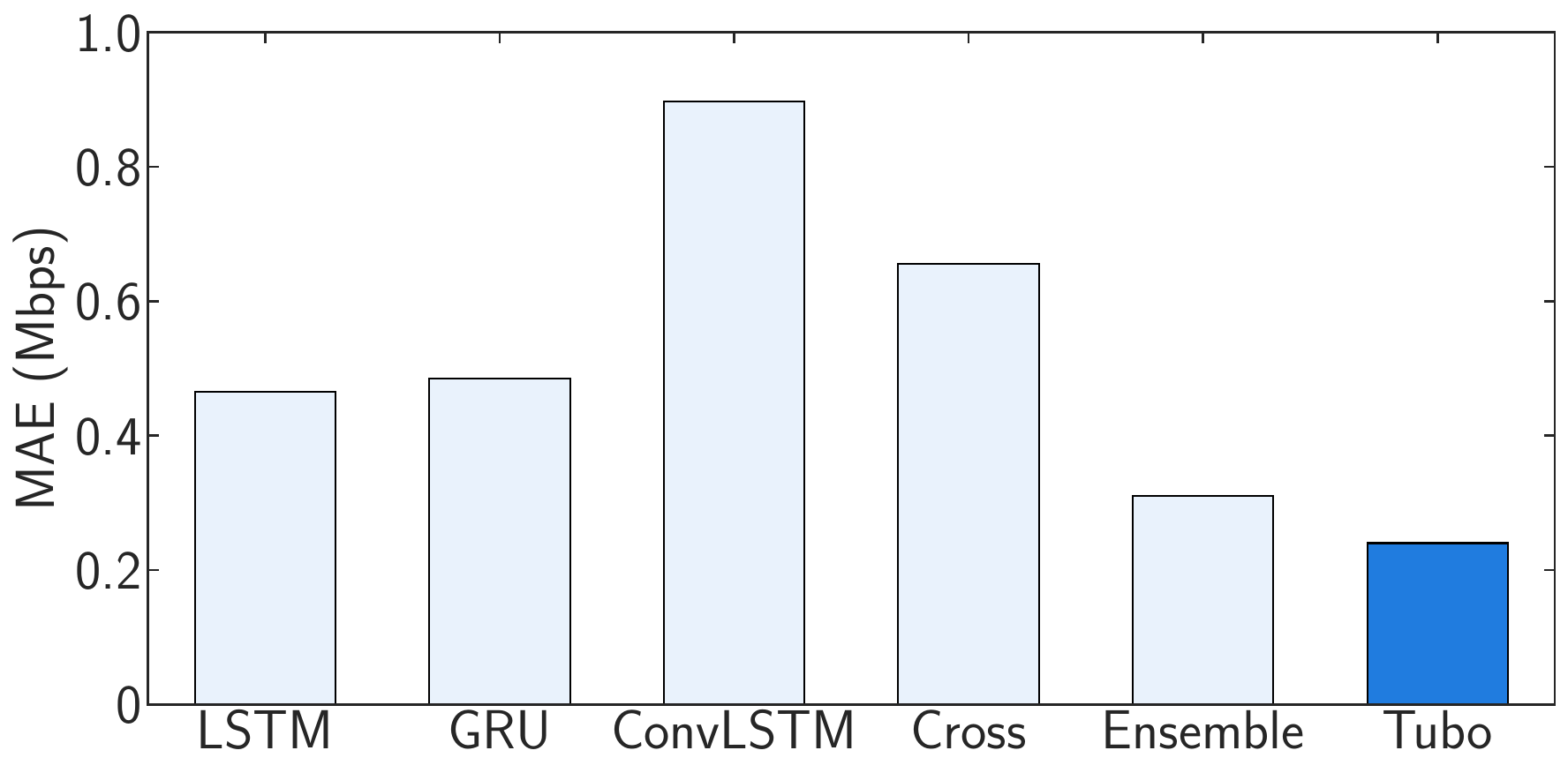}
        \vspace{-0.2in}
        \caption{Abilene}
        \label{fig:overall-mae-abilene-tmgen}
    \end{subfigure}
    \begin{subfigure}{.32\linewidth}
        \includegraphics[width=\linewidth]{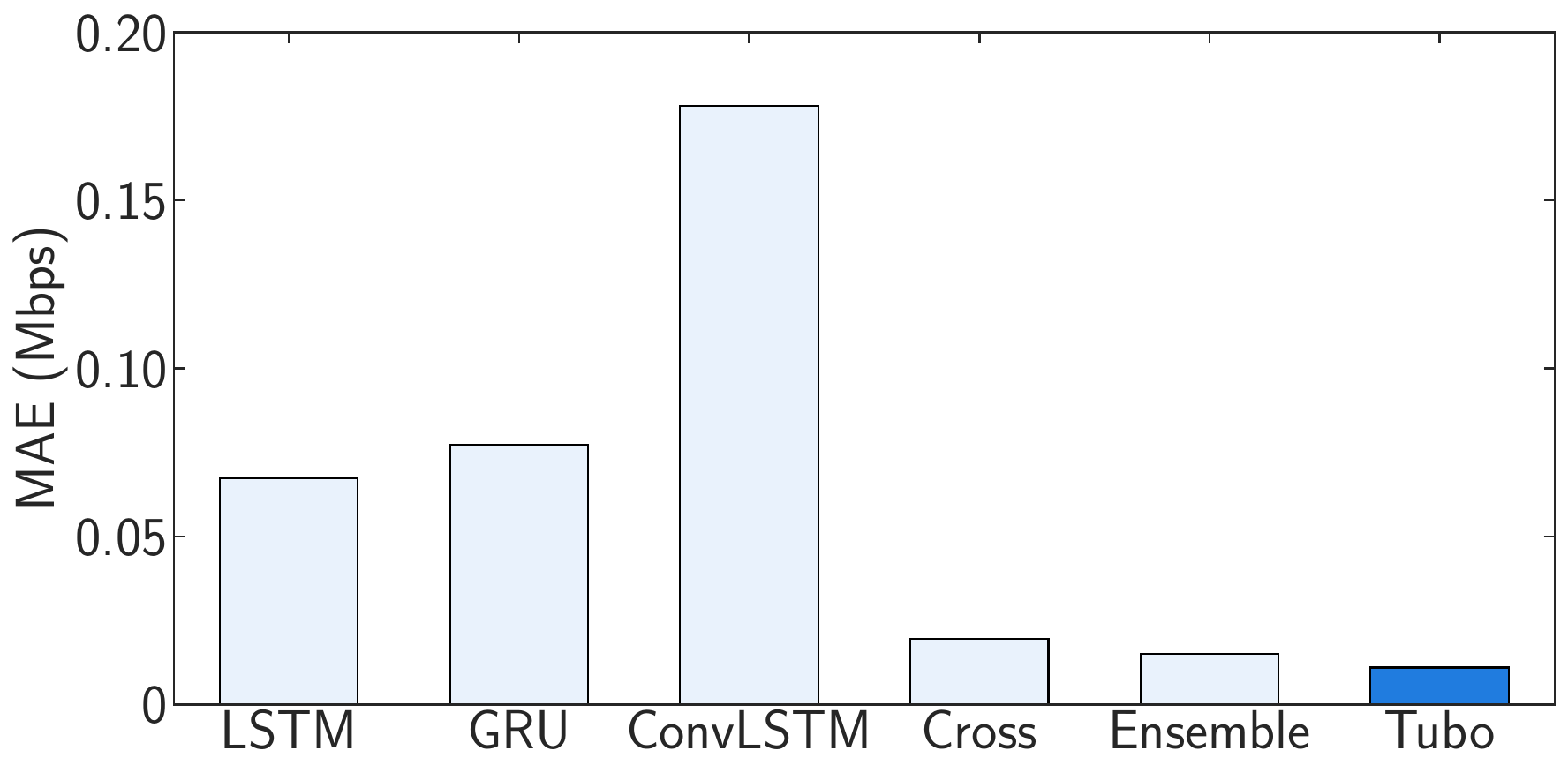}
        \vspace{-0.2in}
        \caption{GEANT}
        \label{fig:overall-mae-geant-tmgen}
    \end{subfigure}
    \begin{subfigure}{.32\linewidth}
        \includegraphics[width=\linewidth]{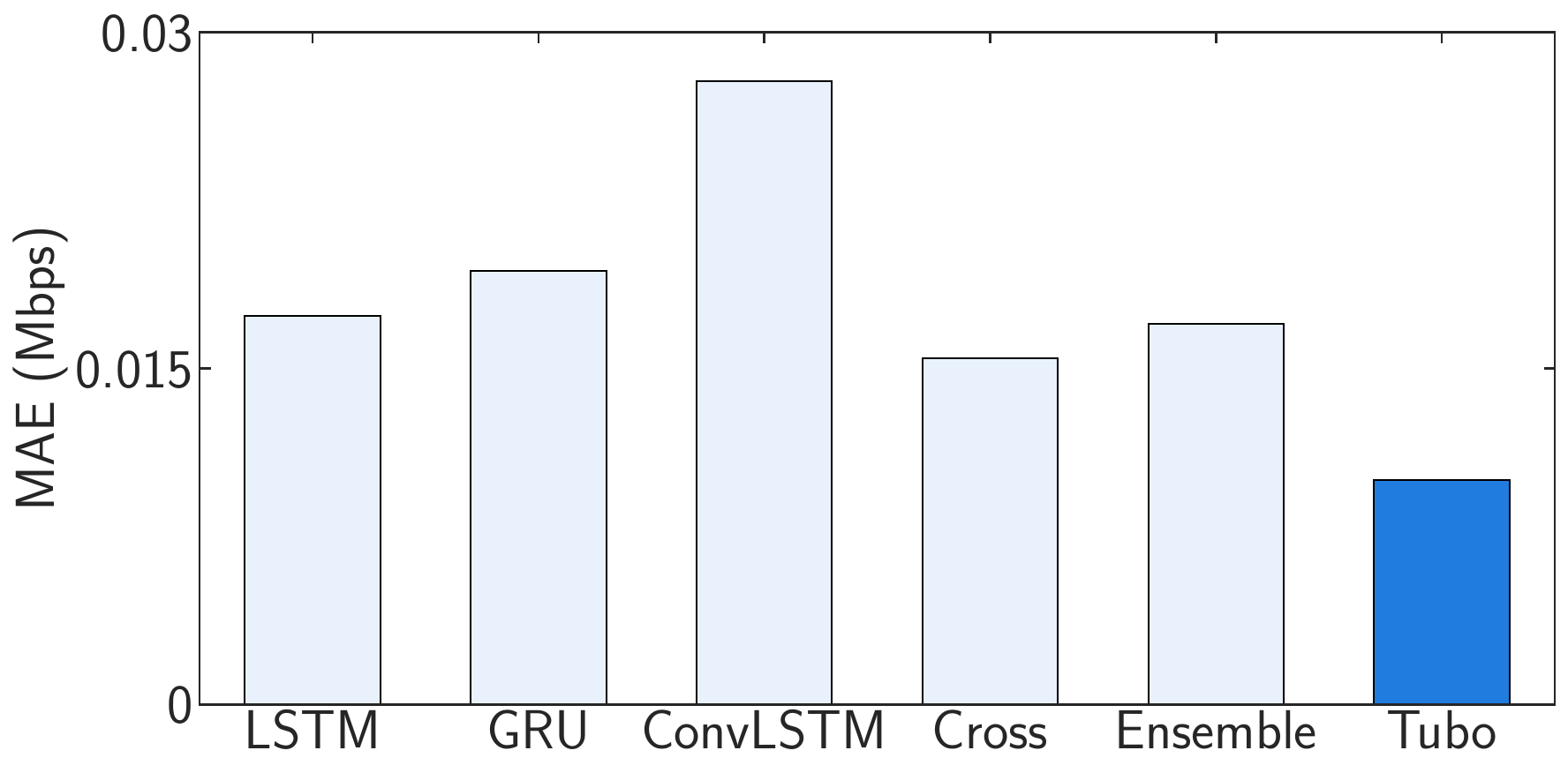}
        \vspace{-0.2in}
        \caption{CERNET}
        \label{fig:overall-mae-cernet-tmgen}
    \end{subfigure}
    \vspace{-0.1in}
    \caption{MAE with synthetic TMGen generated DM data.}
    \label{fig:overall-mae-tmgen}
\end{figure*}

\mypar{MAE with synthetic DM data}
In addition to utilizing real data traffic, we employed  TMGen~\cite{DBLP:conf/sigcomm/TuneR15}, a tool designed for synthetic data traffic generation, to create synthetic datasets for the Abilene, GEANT and CERNET topologies. We specifically opted for the \emph{Modulated Gravity TM} model, calibrating it with mean traffic values, and spatial and temporal variances derived from the actual traffic data corresponding to each topology.
We compare the MAE performance of \sys against various individual forecasting models with synthetic DM data and show the results in Figure~\ref{fig:overall-mae-tmgen}.
We observe that the error is much smaller on this synthetic DM dataset compared to real-world DM datasets considered earlier -- MAE of 0.9 Mbps with \sys in Abilene versus 0.24 Mbps with synthetic data. 
This lower error can be attributed to the more predictable nature of synthetic data, as opposed to real traffic data, which is often influenced by numerous stochastic events that complicate prediction. Even with this relatively less challenging synthetic DM dataset, \sys\ is seen to perform better than any single model. Moreover, even for synthetic data, \sys\ continues to outperform ensemble learning methods, demonstrating consistent superiority as observed with real-world data.

\begin{figure*}[t]
    \centering    
    \begin{subfigure}{.32\linewidth}
        \includegraphics[width=\linewidth]{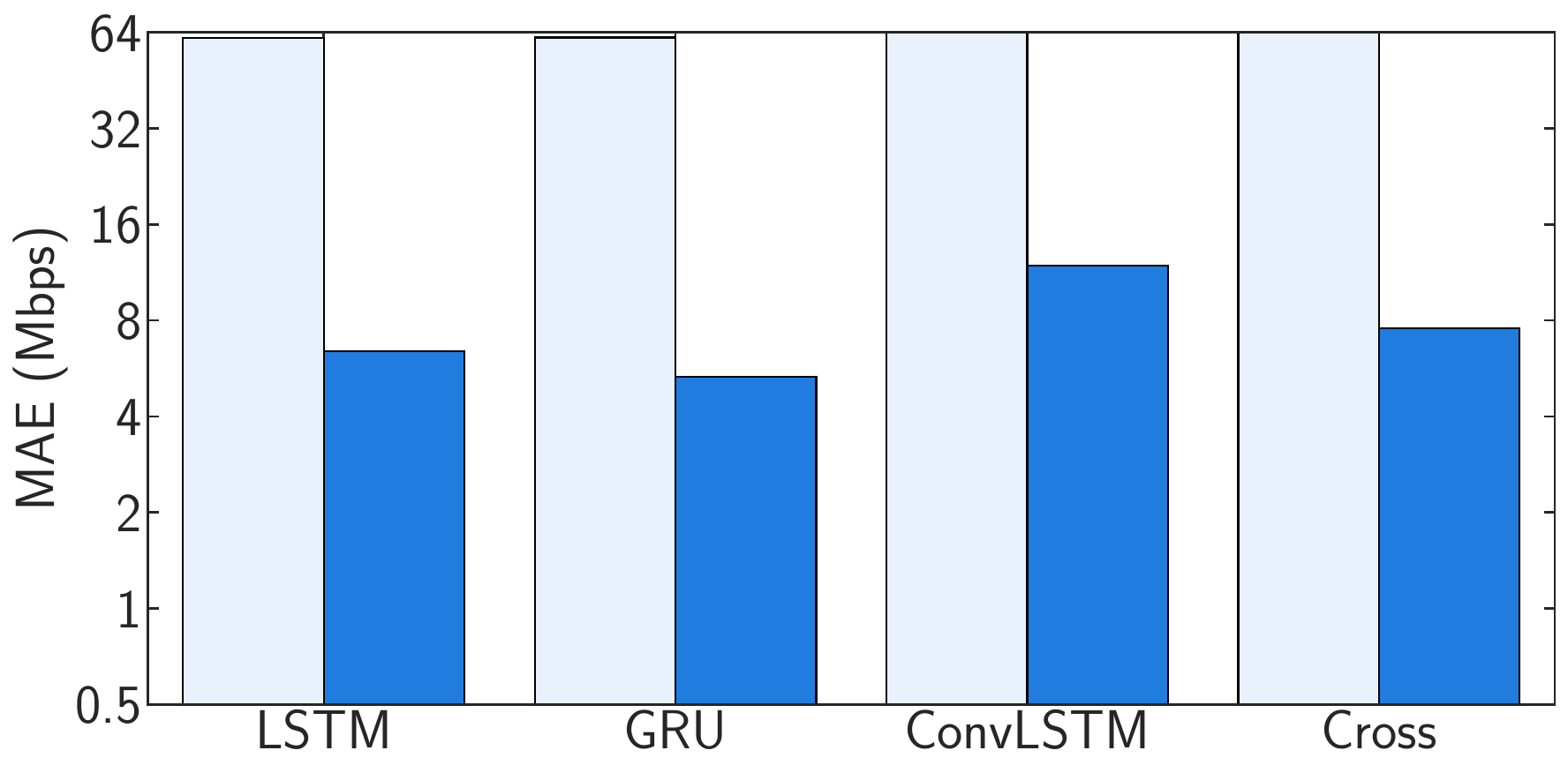}
        \vspace{-0.2in}
        \caption{Abilene}
        \label{fig:overall-burst-abilene}
    \end{subfigure}
    \begin{subfigure}{.32\linewidth}
        \includegraphics[width=\linewidth]{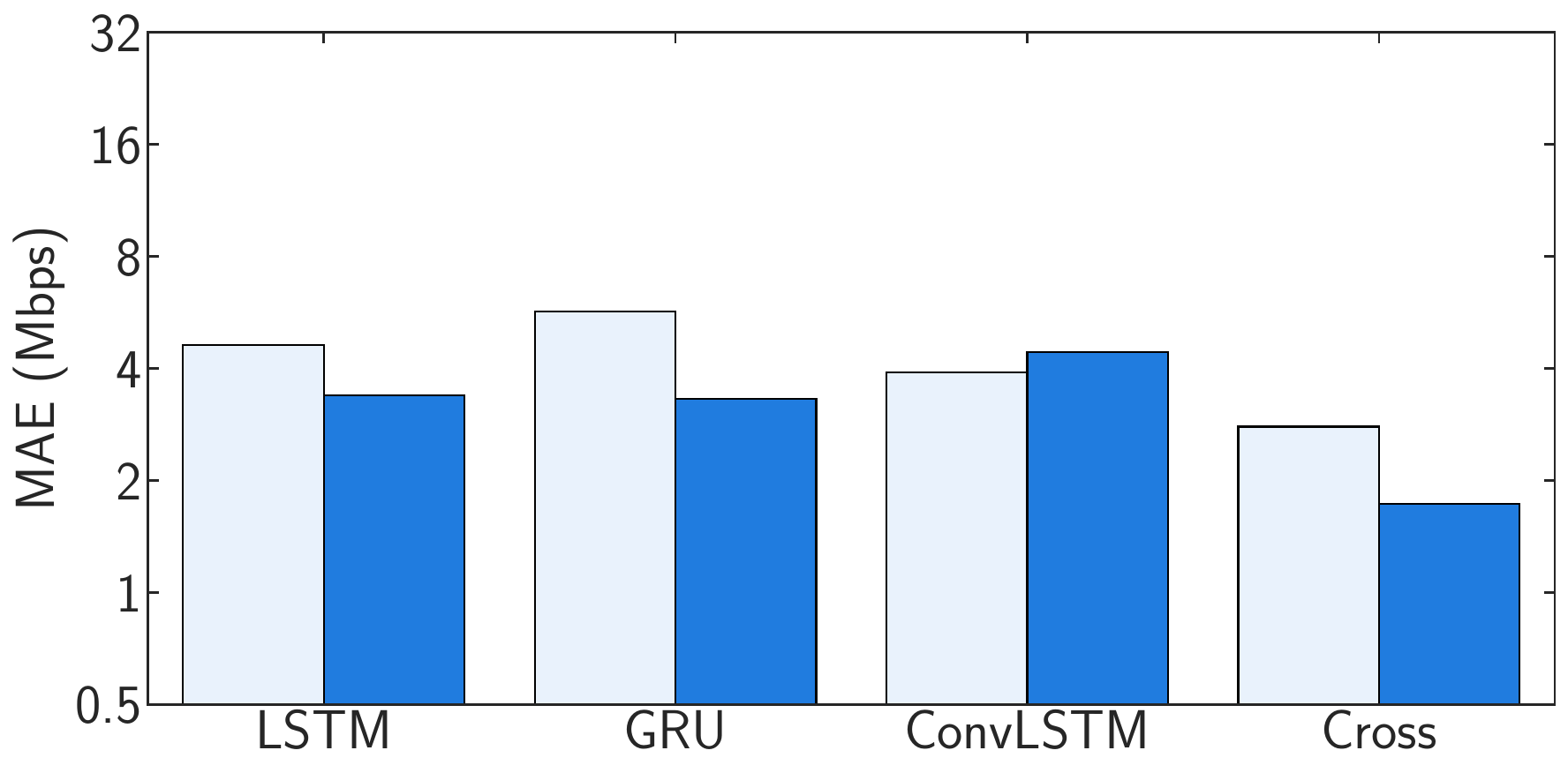}
        \vspace{-0.2in}
        \caption{GEANT}
        \label{fig:overall-burst-geant}
    \end{subfigure}
    \begin{subfigure}{.32\linewidth}
        \includegraphics[width=\linewidth]{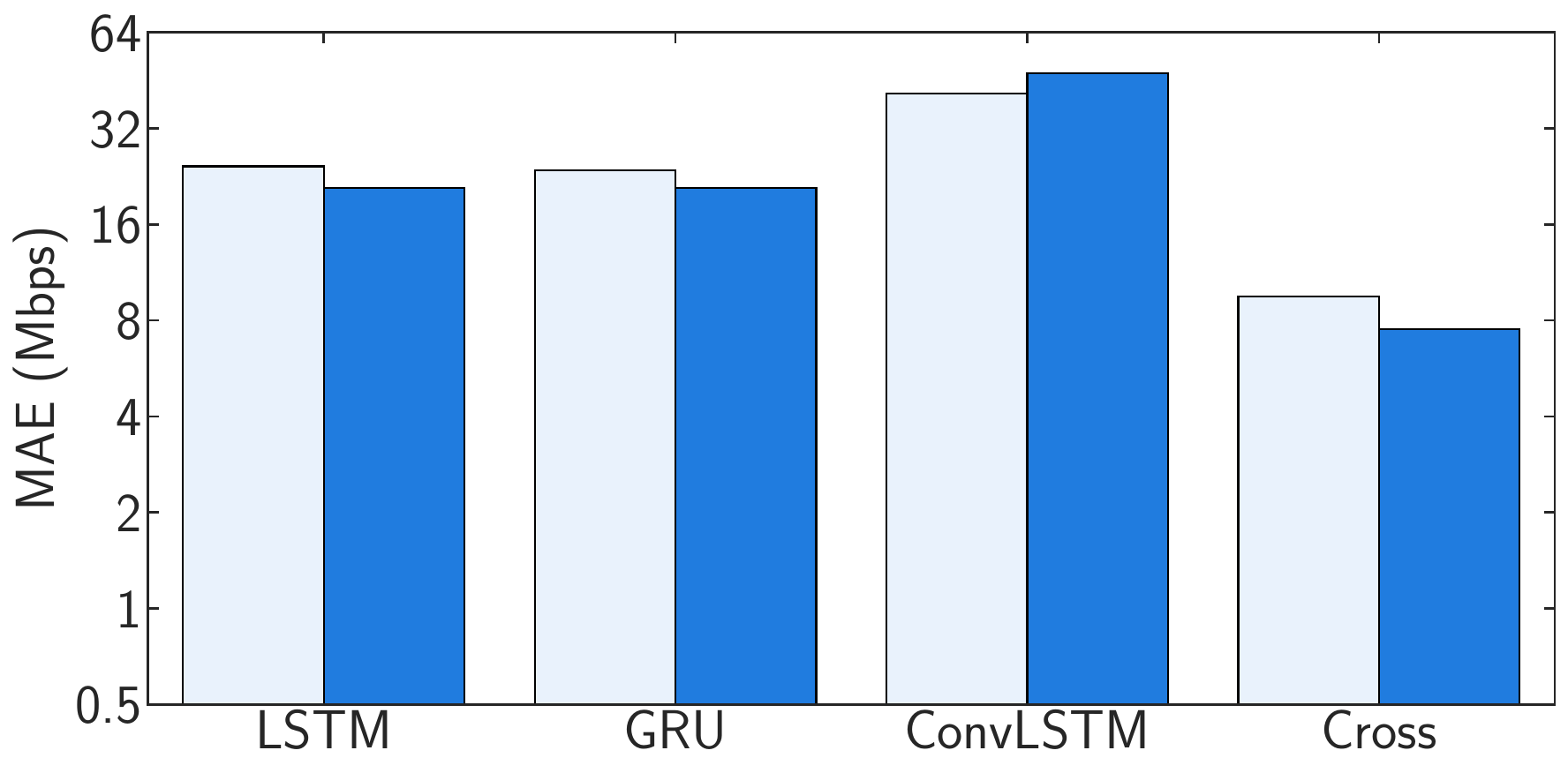}
        \vspace{-0.2in}
        \caption{CERNET}
        \label{fig:overall-burst-cernet}
    \end{subfigure}
    \caption{Impact of burst clipping on MAE. Light-coloured bars show forecasting using raw data, and dark-coloured bars show forecasting with burst clipped.}
    \label{fig:overall-clip}
\end{figure*}

\subsection{TUBO Ablation Study}\label{sec:evaluation-micro}

In this section, we present the performance breakdown of each component within \sys. 
We evaluate the following to understand their contributions to the forecasting accuracy: 
(i) the effectiveness of burst clipping,
(ii) the effectiveness of normalization,
(iii) the accuracy of burst forecasting,
(iv) the effectiveness of model selection.

\mypar{Impact of burst clipping}
In Fig.~\ref{fig:overall-burst-abilene},~\ref{fig:overall-burst-geant},~\ref{fig:overall-burst-cernet}, by comparing training with (w/) and without (w/o) removing bursts, we observe that the extreme values (bursts) do affect the accuracy of forecasting.
Although burst only takes up less than 5\% of the data, depending on its absolute value, we can achieve up to 8$\times$ MAE improvement (\eg LSTM in Abilene from 64 Mbps down to 8 Mbps).
This method is most beneficial for Abilene since it has the largest burst (as shown in Table~\ref{tab:burst-analysis}).
This also indicates that the absolute values of bursts, by themselves, are hard to forecast.
The traffic in CERNET dataset is very regular and smooth.
In such a case, the burst clipping algorithm does not have improvement but also does not introduce degradation to forecasting accuracy.

\begin{figure*}[t]
    \centering
    \begin{subfigure}{\linewidth}
        \centering
        \includegraphics[width=.25\linewidth]{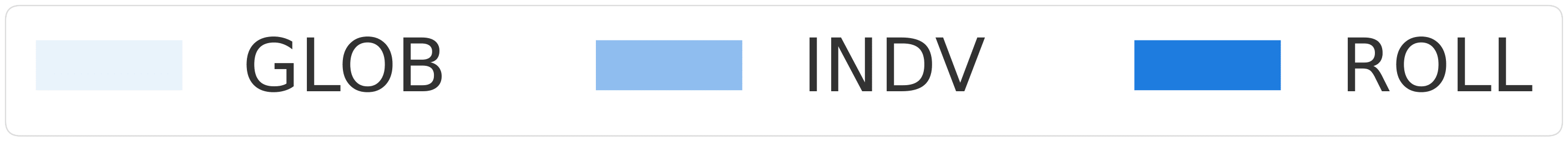}
    \end{subfigure}
    \begin{subfigure}{.32\linewidth}
        \includegraphics[width=\linewidth]{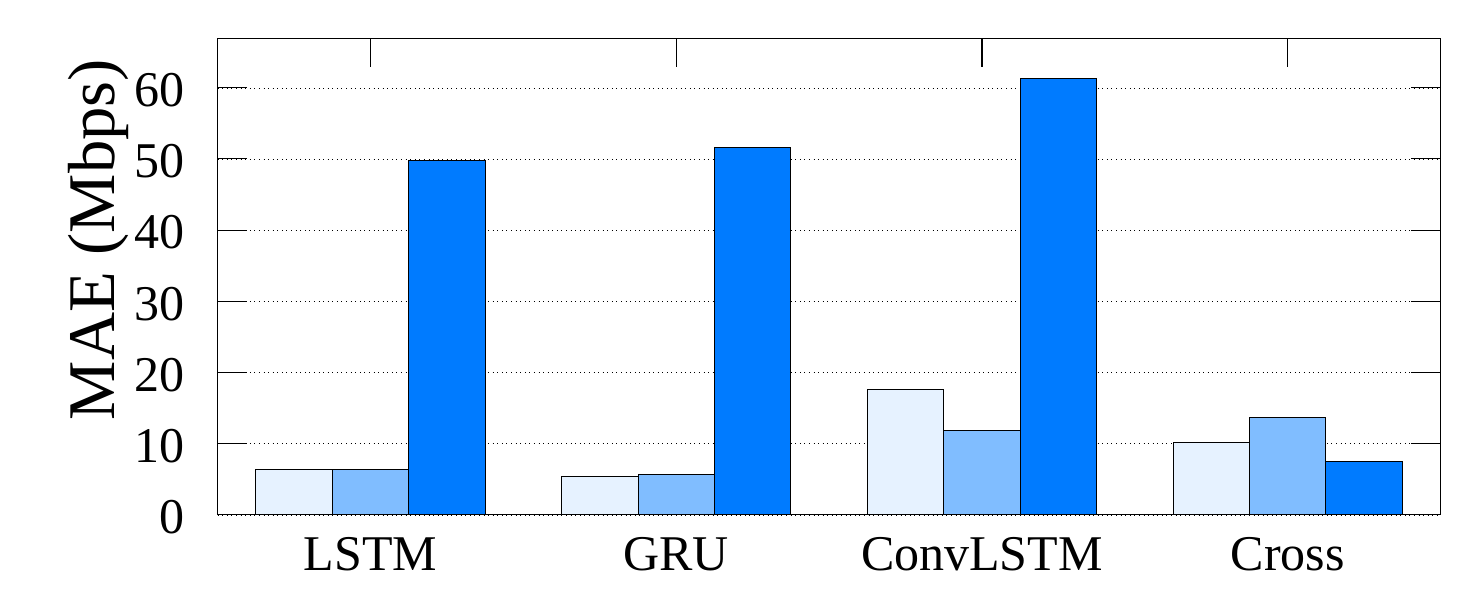}
        \caption{Abilene MAE}
        \label{fig:norm-mae-abilene}
    \end{subfigure}
    \begin{subfigure}{.32\linewidth}
        \includegraphics[width=\linewidth]{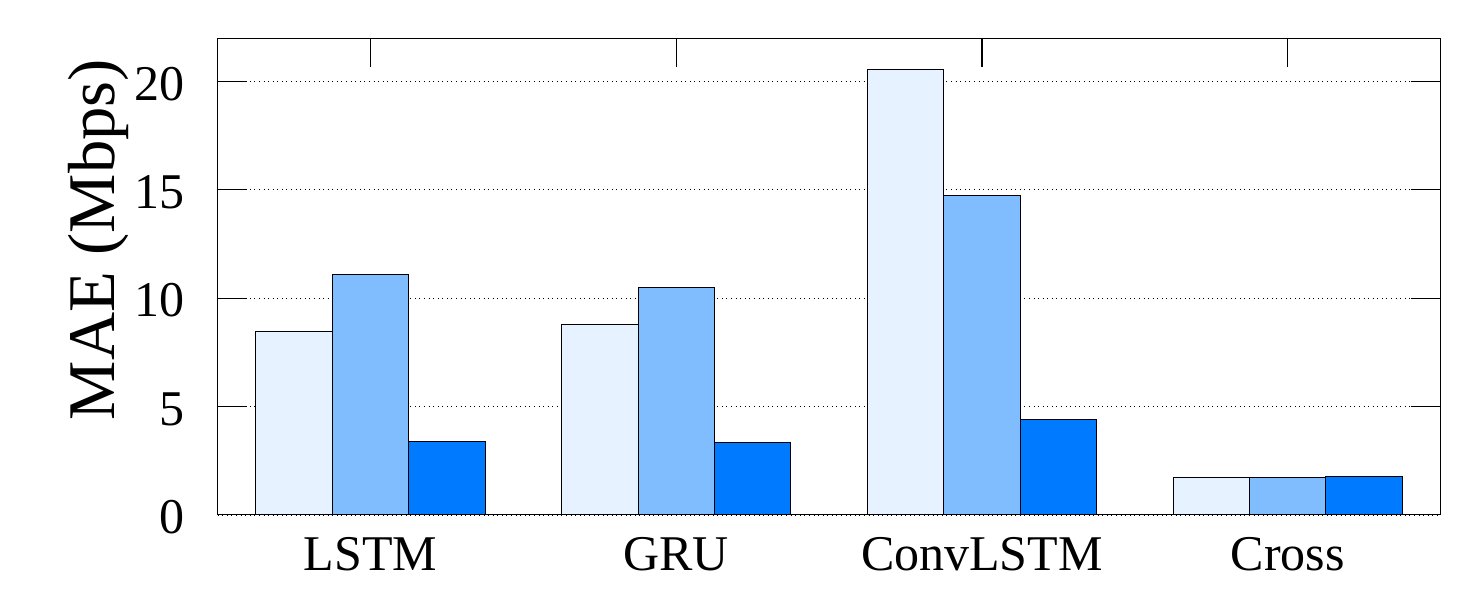}
        \caption{GEANT MAE}
        \label{fig:norm-mae-geant}
    \end{subfigure}
    \begin{subfigure}{.32\linewidth}
        \includegraphics[width=\linewidth]{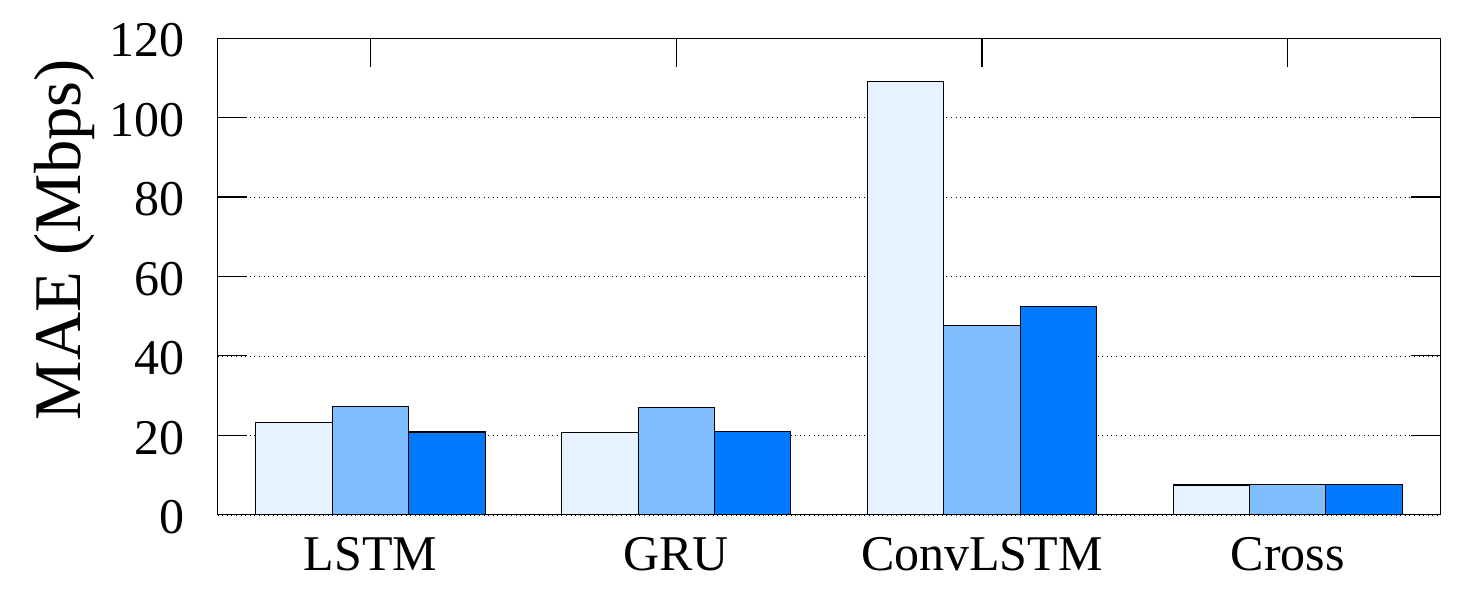}
        \caption{CERNET MAE}
        \label{fig:norm-mae-cernet}
    \end{subfigure}
    \caption{Impact of normalization techniques on MAE with different forecasting methods and DM datasets.}
    \label{fig:norm-mae}
\end{figure*}  

\mypar{Impact of normalization}
Our analysis of normalization techniques underscores their critical role in model training. Mean Absolute Error (MAE) for the various combinations of models and normalization methods is illustrated in Fig.~\ref{fig:norm-mae}.
The primary insight drawn from our exploration is that there is no one-size-fits-all normalization method that performs consistently across all contexts. For instance, in the case of the Abilene dataset, the ROLL method appears well-suited for the Crossformer model. However, this same method does not yield optimal results for other models, likely due to the fluctuating nature of traffic demand.
Furthermore, in the context of the CERNET dataset, we observe minimal performance differences between the ROLL, INDV, and GLOB methods when applied to Crossformer, LSTM, and GRU models. This is attributed to the similarity in traffic patterns across various OD pairs. This emphasizes the necessity of method flexibility and model adaptability in response to specific dataset patterns.

\mypar{Distribution calibration} To evaluate the effectiveness of distribution calibration within \sys, we calculated the Pearson correlation coefficients between MAE and predictive $\sigma$ before and after applying distribution calibration, as described in Section \ref{sec:framework-selection}. The results, shown in Table~\ref{tab:distribution-calibration}, highlight the improvements in correlation for different models (LSTM, GRU, ConvLSTM, and Crossformer) across various datasets (Abilene, GEANT, and CERNET).
The data in Table \ref{tab:distribution-calibration} demonstrate that distribution calibration enhances the correlation between MAE and predictive $\sigma$ across all datasets and model types. For instance, the Crossformer model shows the most significant improvement in the GEANT dataset, with a correlation coefficient of 0.62 post-calibration, compared to 0.43 pre-calibration. These results confirm that distribution calibration effectively aligns predictive uncertainty with true forecasting errors, thereby improving the robustness and reliability of \sys.

Pearson correlation coefficient measures the linear relationship between two variables. Coefficients between 0.5 to 0.6 indicates the predictive uncertainty ($\sigma$) is reasonably aligned with forecasting error (MAE).
In the context of distribution calibration, such a correlation implies that the calibrated uncertainty is a useful predictor of forecasting accuracy.

\begin{table}[t]
    \centering
    \small
    \begin{tabular}{|c|c|c|c|c|}
       \hline
       Dataset  & LSTM       & GRU        & ConvLSTM   & Crossformer \\
       \hline
       Abilene  & 0.51(0.40) & 0.52(0.41) & 0.50(0.39) & 0.60(0.42)  \\
       GEANT    & 0.57(0.41) & 0.52(0.39) & 0.56(0.34) & 0.62(0.43)  \\
       CERNET   & 0.51(0.29) & 0.53(0.38) & 0.53(0.42) & 0.61(0.39)  \\
        \hline
    \end{tabular}
    \caption{Improvement in Pearson correlation coefficients between MAE and predictive $\sigma$ after distribution calibration. Numbers in brackets indicate the correlation coefficients before distribution calibration.}
    \label{tab:distribution-calibration}
\end{table}

\mypar{Model selection algorithm}
To verify the effectiveness of our model selection algorithm within \sys, we conduct an extensive evaluation that considers a variety of parameters. We examine the MAE across all datasets for each OD pair, comparing the performance when using our model selection algorithm versus the performance when using the best model, in this case, the Crossformer. This comparison offers insights into the added value of our model selection algorithm over a static selection of the best model.
In addition to accuracy, we assess the robustness of our model selection algorithm. We do this by applying it in two distinct scenarios: forecasting with the burst component of traffic removed and forecasting without removing the burst. Bursty traffic can significantly affect the performance of forecasting models, so this comparison helps us understand how well our model selection algorithm can adapt to varying levels of traffic dynamics.

\begin{figure}[t]
    \centering
    \begin{subfigure}{0.48\linewidth}
        \includegraphics[width=\linewidth]{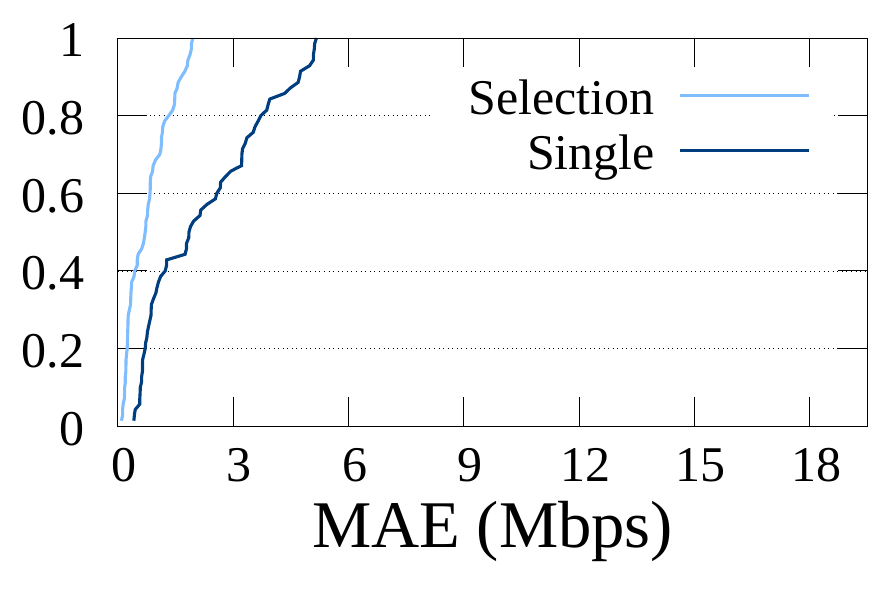}
        \label{fig:model-selection-clip}
    \end{subfigure}
    \begin{subfigure}{0.48\linewidth}
        \includegraphics[width=\linewidth]{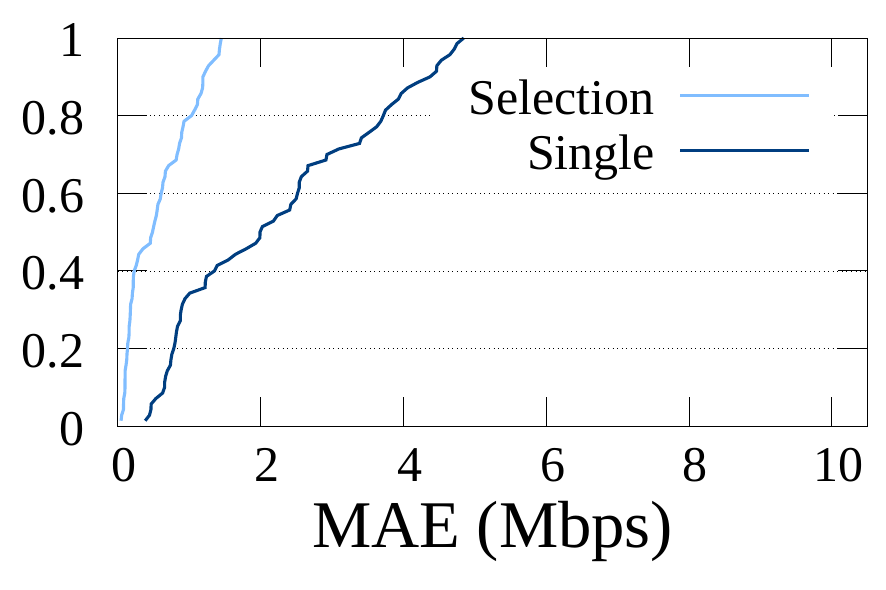}
        \label{fig:model-selection-noclip}
    \end{subfigure}
    \caption{OD pair MAE CDF. Left w/ burst, right w/o burst.}
    \label{fig:model-selection}
\end{figure}

As depicted in Fig.~\ref{fig:model-selection}, our model selection process significantly enhances forecasting accuracy. 
Remarkably, it can improve the average MAE across all OD pairs by up to 2$\times$, and for the worst-case MAE of single OD pairs, the improvement can be as high as 4$\times$.
In terms of tail MAE improvement, we noticed that a single-model forecasting approach can often falter on certain inputs. However, different models, thanks to their unique optimizers and loss functions, do not necessarily fail on the same inputs. This variability in model performance underscores the value of our model selection process, which dynamically chooses the optimal model for each input.

An additional observation from our study is that the performance of the model selection algorithm improves when burst traffic is removed from the data. In particular, removing burst traffic results in a 2$\times$ larger gain in tail forecasting accuracy.
Burst traffic can introduce significant irregularities in the input data used for model training. This irregularity effectively violates the ideal condition that all input data should be on the same scale.
This observation underscores the critical role of data preprocessing with bursts handled separately to improve the overall performance of proactive traffic engineering.

Table~\ref{tab:model-select} illustrates the diversity of model selection in our pipeline. Although Crossformer shows the best overall accuracy, it is not universally the optimal choice for every input. 
While the algorithm often selects Crossformer, it also considers other models that, despite a slightly higher MAE, can perform better in specific contexts. 
These results reinforce the principle that a single model cannot optimally handle all scenarios.
Different traffic patterns and demand dynamics might be better handled by specific models. For instance, an LSTM model might excel at capturing long-term dependencies in the data, whereas ConvLSTM might perform better in scenarios where localized features are more relevant. 
This demonstrates the strength of our approach, where the ability to select and deploy the most suitable model for each specific input can significantly enhance overall forecasting accuracy. 

\begin{table}[t]
    \centering
    \small
    \begin{tabular}{|c|c|c|c|c|}
       \hline
       Dataset  & LSTM & GRU    & ConvLSTM & Cross  \\
       \hline
       Abilene  & 15\% & 17\%   &  20\%    &  49\%  \\
       GEANT    & 18\% & 19\%   &  15\%    &  48\%  \\
       CERNET   & 10\% & 14\%   &  17\%    &  59\%  \\
        \hline
    \end{tabular}
    \caption{Model selection ratio on evaluation datasets.}
    \label{tab:model-select}
\end{table}

\mypar{Burst occurrence forecasting}
Our evaluation extends to testing the accuracy of burst detection across all three datasets.
The burst forecasting model is implemented using the Crossformer~\cite{DBLP:conf/iclr/0001YCL00L22} and trained with the L1 loss function. The evaluation of burst occurrence is twofold, as summarized in Table~\ref{tab:burst-acc}. First, we measure the forecasting accuracy, which evaluates whether bursts are correctly classified. Given the scarcity of burst occurrences, we further assess the performance using the Burst Identification Rate, which quantifies the proportion of actual bursts that are accurately forecasted.
From Table~\ref{tab:burst-acc}, the highest overall accuracy is 93.89\%, observed with the Abilene dataset, while the lowest is 88.86\% with CERNET. Similarly, the Burst Identification Rate is highest for GEANT at 94.35\% and lowest for CERNET at 86.32\%. The consistently high overall accuracy and Burst Identification Rates indicate that burst occurrences, irrespective of their magnitude, exhibit a significant degree of structure and correlation across all OD pairs.
This realization provides us with an essential understanding that even though burst traffic can introduce irregularities in the data, these bursts are not entirely random and their occurrence can be accurately predicted. This offers another layer of sophistication when applied to the downstream use case of proactive traffic engineering by providing the potential to anticipate and effectively manage burst traffic.

\begin{table}[]
\centering
\small
\begin{tabular}{|c|c|c|}
\hline
Dataset & \begin{tabular}[c]{@{}c@{}}Overall \\ (Burst Percentage)\end{tabular} & \begin{tabular}[c]{@{}c@{}}Burst Identification \\ Rate\end{tabular} \\ \hline
Abilene & 93.89\% (3.37\%) & 92.51\% \\
GEANT   & 92.98\% (2.92\%) & 94.35\% \\
CERNET  & 88.86\% (1.57\%) & 86.32\% \\ \hline
\end{tabular}
\caption{Burst forecasting accuracy on evaluation datasets.}
   \label{tab:burst-acc}
\end{table}

\subsection{Use Case: Proactive Traffic Engineering}\label{sec:evaluation-e2e}

Here we showcase the benefit of DM forecasting with \sys for enabling efficient and {\em proactive} TE. We consider aggregate throughput, measured in Megabits per second (Mbps), as the metric. 
We evaluate two TE objectives: max-commodity-flow ($p_1$) and low-latency maximum flow ($p_2$)~\cite{DBLP:conf/sosr/SinghBK22}.
We choose reactive oblivious routing TE (SMORE~\cite{DBLP:conf/nsdi/KumarYYFKLLS18}) as one baseline and proactive TE with ConvLSTM~\cite{jiang2022internet} (SOTA DM forecasting method) as a second baseline.
As for SMORE, each OD pair chose one shortest path as the tunnel.
We normalize the throughput according to the output of linear programming using the ground truth.
For each epoch in our dataset, we calculate the result of implementing proactive TE across the network, which we then apply to the succeeding epoch.
We omitted the LSTM results for the CERNET dataset due to LSTM's tendency to overestimate traffic demand values, resulting in unfeasible TE strategies.

\begin{figure}[t]
    \centering
    \begin{subfigure}{0.9\linewidth}
        \includegraphics[width=\linewidth]{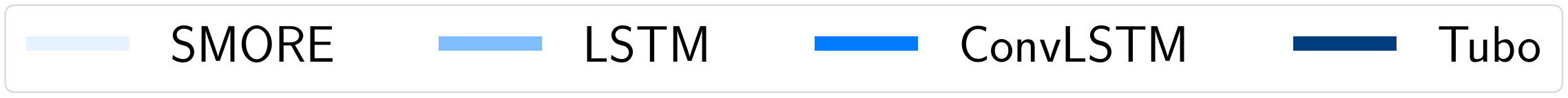}
    \end{subfigure}
    \begin{subfigure}{0.99\linewidth}
        \includegraphics[width=\linewidth]{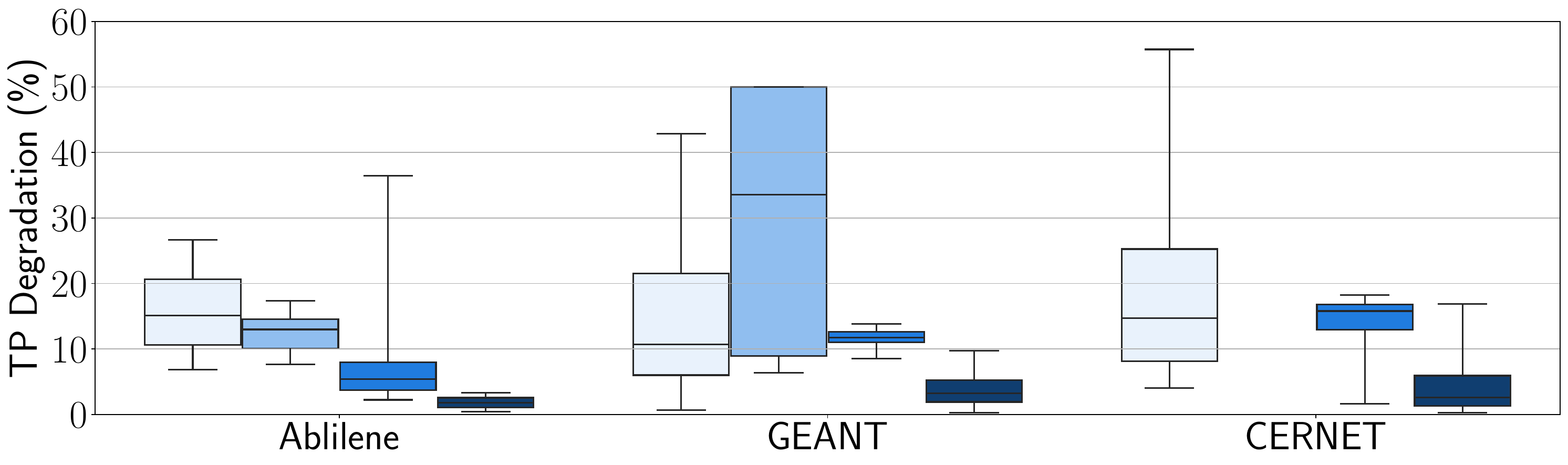}
        \label{fig:eval-te-abilene-p1}
    \end{subfigure}
    \begin{subfigure}{0.99\linewidth}
        \includegraphics[width=\linewidth]{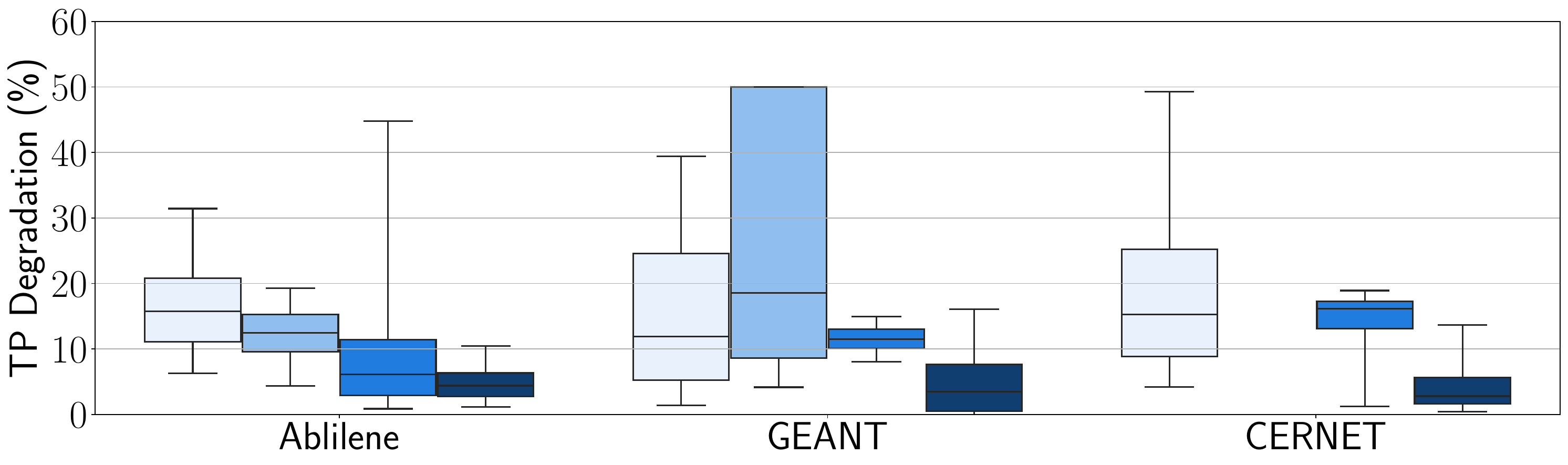}
        \label{fig:eval-te-abilene-p2}
    \end{subfigure}
    \caption{TE result $p_1$ (top) and $p_2$ (bottom). In terms of throughput (TP) degradation, the lower values mean closer to ideal throughput. The whiskers show the 5 and 95 percentile, indicating the worst and best performance.}
    \label{fig:eval-te}
\end{figure}

The results from Fig.~\ref{fig:eval-te} demonstrate that proactive TE significantly outperforms its reactive counterpart in terms of throughput, while the process of model selection effectively mitigates errors when compared to using a single state-of-the-art (SOTA) DM forecasting model, specifically ConvLSTM~\cite{jiang2022internet}.
In the context of the $p1$ problem, \sys significantly improves performance, reducing the median throughput penalty by up to 9$\times$ (from 16\% to 1.7\%) when compared to SMORE in Abilene dataset. This represents a significant enhancement even over the SOTA DM forecasting model, with \sys reducing the median penalty by up to 3$\times$ (from 5.22\% to 1.7\%).
For the $p2$ problem, \sys continues to exhibit superior performance. It improves the situation by up to 4$\times$ (from 16.64\% to 4.01\%) compared to reactive techniques and also shows a reduction of up to 1.63$\times$ (from 6.57\% to 4.01\%) when compared to the state-of-the-art DM forecasting model.
These results underscore the substantial improvements in traffic engineering performance that \sys can deliver.
The result demonstrates that when selecting forecasting models according to TE performance, we can transfer the benefits from DM forecasting to TE objectives.
As for baselines, the forecasting objective does not have a correlation with TE; although the MAE performance might be quite close, the delivery of TE throughput is not aligned.
This underscores the significant role that the intelligent selection of a suitable forecasting model can play in enhancing the overall performance and precision of TE, emphasizing the advantages offered by \sys.

\section{Conclusions}

In this work, we present \sys, an innovative machine learning (ML) framework tailored for robust network demand matrix (DM) forecasting. 
The core components of \sys, including burst processing and model selection, enable it to effectively manage bursts and adapt to varying DM patterns, latter via uncertainty driven model selection.
Our comprehensive evaluations, conducted on real-world network demand matrix datasets including Abilene, GEANT, and CERNET, show significant improvement in DM forecasting accuracy compared to existing methods by up to 4$\times$. 
Furthermore, by providing reliable forecasted DMs in presence of variability and bursts in network traffic demands, \sys can benefit downstream traffic forecasting based applications. For the TE downstream use case, \sys is shown to 
yield significant gains in TE performance up to 9$\times$ over traditional oblivious routing approach and up to 3$\times$ over
proactive TE with the state-of-the-art DM forecasting method (ConvLSTM). 

\bibliographystyle{IEEEtran}
\bibliography{arxiv}

@String{Computing = "Computing" }

@String{Computer = "{IEEE} Computer" }

@inproceedings{DBLP:conf/infocom/GaoLCGGCG20,
  author    = {Kaihui Gao and
               Dan Li and
               Li Chen and
               Jinkun Geng and
               Fei Gui and
               Yang Cheng and
               Yue Gu},
  title     = {Predicting Traffic Demand Matrix by Considering Inter-flow Correlations},
  booktitle = {{INFOCOM} Workshops},
  pages     = {165--170},
  publisher = {{IEEE}},
  year      = {2020}
}

@inproceedings{DBLP:conf/iclr/LiYS018,
  author       = {Yaguang Li and
                  Rose Yu and
                  Cyrus Shahabi and
                  Yan Liu},
  title        = {Diffusion Convolutional Recurrent Neural Network: Data-Driven Traffic
                  Forecasting},
  booktitle    = {{ICLR} (Poster)},
  publisher    = {OpenReview.net},
  year         = {2018}
}

@article{DBLP:journals/gpem/Heaton18,
  author       = {Jeff Heaton},
  title        = {Ian Goodfellow, Yoshua Bengio, and Aaron Courville: Deep learning
                  - The {MIT} Press, 2016, 800 pp, {ISBN:} 0262035618},
  journal      = {Genet. Program. Evolvable Mach.},
  volume       = {19},
  number       = {1-2},
  pages        = {305--307},
  year         = {2018}
}

@inproceedings{10.1145/1015330.1015435,
author = {Ng, Andrew Y.},
title = {Feature Selection, L1 vs. L2 Regularization, and Rotational Invariance},
year = {2004},
isbn = {1581138385},
publisher = {Association for Computing Machinery},
address = {New York, NY, USA},
booktitle = {ICML},
}

@inproceedings{adam,
  author       = {Diederik P. Kingma and
                  Jimmy Ba},
  title        = {Adam: {A} Method for Stochastic Optimization},
  booktitle    = {{ICLR} (Poster)},
  year         = {2015}
}

@book{box2015time,
  title={Time series analysis: forecasting and control},
  author={Box, George EP and Jenkins, Gwilym M and Reinsel, Gregory C and Ljung, Greta M},
  year={2015},
  publisher={John Wiley \& Sons}
}

@inproceedings{DBLP:conf/cnsm/LazarisP19,
  author    = {Aggelos Lazaris and
               Viktor K. Prasanna},
  title     = {Deep Learning Models For Aggregated Network Traffic Prediction},
  booktitle = {{CNSM}},
  pages     = {1--5},
  publisher = {{IEEE}},
  year      = {2019}
}

@inproceedings{DBLP:conf/emnlp/ChoMGBBSB14,
  author       = {Kyunghyun Cho and
                  Bart van Merrienboer and
                  {\c{C}}aglar G{\"{u}}l{\c{c}}ehre and
                  Dzmitry Bahdanau and
                  Fethi Bougares and
                  Holger Schwenk and
                  Yoshua Bengio},
  title        = {Learning Phrase Representations using {RNN} Encoder-Decoder for Statistical
                  Machine Translation},
  booktitle    = {{EMNLP}},
  pages        = {1724--1734},
  publisher    = {{ACL}},
  year         = {2014}
}

@article{DBLP:journals/neco/HochreiterS97,
  author    = {Sepp Hochreiter and
               J{\"{u}}rgen Schmidhuber},
  title     = {Long Short-Term Memory},
  journal   = {Neural Comput.},
  volume    = {9},
  number    = {8},
  pages     = {1735--1780},
  year      = {1997}
}

@inproceedings{DBLP:conf/nsdi/KumarYYFKLLS18,
  author    = {Praveen Kumar and
               Yang Yuan and
               Chris Yu and
               Nate Foster and
               Robert Kleinberg and
               Petr Lapukhov and
               Chiunlin Lim and
               Robert Soul{\'{e}}},
  title     = {Semi-Oblivious Traffic Engineering: The Road Not Taken},
  booktitle = {{NSDI}},
  pages     = {157--170},
  publisher = {{USENIX} Association},
  year      = {2018}
}

@article{hochreiter1997long,
  title={Long short-term memory},
  author={Hochreiter, Sepp and Schmidhuber, J{\"u}rgen},
  journal={Neural computation},
  volume={9},
  number={8},
  pages={1735--1780},
  year={1997},
  publisher={MIT Press}
}

@article{schuster1997bidirectional,
  title={Bidirectional recurrent neural networks},
  author={Schuster, Mike and Paliwal, Kuldip K},
  journal={IEEE transactions on Signal Processing},
  volume={45},
  number={11},
  pages={2673--2681},
  year={1997},
  publisher={Ieee}
}

@inproceedings{DBLP:conf/sosr/SinghBK22,
  author    = {Rachee Singh and
               Nikolaj S. Bj{\o}rner and
               Umesh Krishnaswamy},
  title     = {Traffic engineering: from {ISP} to cloud wide area networks},
  booktitle = {{SOSR}},
  pages     = {50--58},
  publisher = {{ACM}},
  year      = {2022}
}

@inproceedings{DBLP:conf/iscc/LiuWYSGT19,
  author    = {Zhifeng Liu and
               Zhiliang Wang and
               Xia Yin and
               Xingang Shi and
               Yingya Guo and
               Ying Tian},
  title     = {Traffic Matrix Prediction Based on Deep Learning for Dynamic Traffic
               Engineering},
  booktitle = {{ISCC}},
  pages     = {1--7},
  publisher = {{IEEE}},
  year      = {2019}
}

@inproceedings{DBLP:conf/sigmetrics/ZhangRDG03,
  author    = {Yin Zhang and
               Matthew Roughan and
               Nick G. Duffield and
               Albert G. Greenberg},
  title     = {Fast accurate computation of large-scale {IP} traffic matrices from
               link loads},
  booktitle = {{SIGMETRICS}},
  pages     = {206--217},
  publisher = {{ACM}},
  year      = {2003}
}

@article{DBLP:journals/ccr/UhligQLB06,
  author    = {Steve Uhlig and
               Bruno Quoitin and
               Jean Lepropre and
               Simon Balon},
  title     = {Providing public intradomain traffic matrices to the research community},
  journal   = {Comput. Commun. Rev.},
  volume    = {36},
  number    = {1},
  pages     = {83--86},
  year      = {2006}
}

@inproceedings{DBLP:conf/iclr/0001YCL00L22,
  author       = {Wenxiao Wang and
                  Lu Yao and
                  Long Chen and
                  Binbin Lin and
                  Deng Cai and
                  Xiaofei He and
                  Wei Liu},
  title        = {CrossFormer: {A} Versatile Vision Transformer Hinging on Cross-scale
                  Attention},
  booktitle    = {{ICLR}},
  publisher    = {OpenReview.net},
  year         = {2022}
}

@article{DBLP:journals/corr/FlunkertSG17,
  author    = {Valentin Flunkert and
               David Salinas and
               Jan Gasthaus},
  title     = {{DeepAR}: Probabilistic Forecasting with Autoregressive Recurrent Networks},
  journal   = {CoRR},
  volume    = {abs/1704.04110},
  year      = {2017}
}

@article{de2011forecasting,
  title={Forecasting time series with complex seasonal patterns using exponential smoothing},
  author={De Livera, Alysha M and Hyndman, Rob J and Snyder, Ralph D},
  journal={Journal of the American statistical association},
  volume={106},
  number={496},
  pages={1513--1527},
  year={2011},
  publisher={Taylor \& Francis}
}

@article{grubbs1969procedures,
  title={Procedures for detecting outlying observations in samples},
  author={Grubbs, Frank E},
  journal={Technometrics},
  volume={11},
  number={1},
  pages={1--21},
  year={1969},
  publisher={Taylor \& Francis}
}

@inproceedings{DBLP:conf/cvpr/HeZRS16,
  author       = {Kaiming He and
                  Xiangyu Zhang and
                  Shaoqing Ren and
                  Jian Sun},
  title        = {Deep Residual Learning for Image Recognition},
  booktitle    = {{CVPR}},
  pages        = {770--778},
  publisher    = {{IEEE} Computer Society},
  year         = {2016}
}

@misc{ciscoreport,
  author = {{CISCO}},
  title = {Global - 2021 Forecast Highlights},
  year = {2021},
  howpublished = {\url{https://www.cisco.com/c/dam/m/en_us/solutions/service-provider/vni-forecast-highlights/pdf/Global_2021_Forecast_Highlights.pdf}}
}

@article{jiang2022internet,
  title={Internet traffic matrix prediction with convolutional {LSTM} neural network},
  author={Jiang, Weiwei},
  journal={Internet Technology Letters},
  volume={5},
  number={2},
  pages={e322},
  year={2022},
  publisher={Wiley Online Library}
}

@book{goodfellow2016deep,
  title={Deep Learning},
  author={Goodfellow, Ian and Bengio, Yoshua and Courville, Aaron},
  year={2016},
  publisher={MIT press}
}

@inproceedings{DBLP:conf/sigcomm/TuneR15,
  author       = {Paul Tune and
                  Matthew Roughan},
  title        = {Spatiotemporal Traffic Matrix Synthesis},
  booktitle    = {{SIGCOMM}},
  pages        = {579--592},
  publisher    = {{ACM}},
  year         = {2015}
}

@article{roberts2013gaussian,
  title={Gaussian processes for time-series modelling},
  author={Roberts, Stephen and Osborne, Michael and Ebden, Mark and Reece, Steven and Gibson, Neale and Aigrain, Suzanne},
  journal={Philosophical Transactions of the Royal Society A: Mathematical, Physical and Engineering Sciences},
  volume={371},
  number={1984},
  pages={20110550},
  year={2013},
  publisher={The Royal Society Publishing}
}

@inproceedings{zeng2023transformers,
  title={Are transformers effective for time series forecasting?},
  author={Zeng, Ailing and Chen, Muxi and Zhang, Lei and Xu, Qiang},
  booktitle={Proceedings of the AAAI conference on artificial intelligence},
  volume={37},
  number={9},
  pages={11121--11128},
  year={2023}
}

@inproceedings{song2019distribution,
  title={Distribution calibration for regression},
  author={Song, Hao and Diethe, Tom and Kull, Meelis and Flach, Peter},
  booktitle={International Conference on Machine Learning},
  pages={5897--5906},
  year={2019},
  organization={PMLR}
}

@inproceedings{gal2016dropout,
  title={Dropout as a bayesian approximation: Representing model uncertainty in deep learning},
  author={Gal, Yarin and Ghahramani, Zoubin},
  booktitle={international conference on machine learning},
  pages={1050--1059},
  year={2016},
  organization={PMLR}
}

@article{amini2020deep,
  title={Deep evidential regression},
  author={Amini, Alexander and Schwarting, Wilko and Soleimany, Ava and Rus, Daniela},
  journal={Advances in neural information processing systems},
  volume={33},
  pages={14927--14937},
  year={2020}
}

@article{wassie2023traffic,
  title={Traffic prediction in {SDN} for explainable {QoS} using deep learning approach},
  author={Wassie, Getahun and Ding, Jianguo and Wondie, Yihenew},
  journal={Scientific Reports},
  volume={13},
  number={1},
  pages={20607},
  year={2023},
  publisher={Nature Publishing Group UK London}
}

@article{wu2022timesnet,
  title={Timesnet: Temporal 2d-variation modeling for general time series analysis},
  author={Wu, Haixu and Hu, Tengge and Liu, Yong and Zhou, Hang and Wang, Jianmin and Long, Mingsheng},
  journal={arXiv preprint arXiv:2210.02186},
  year={2022},
  publisher={arXivpreprint}
}

@article{ansari2024chronos,
  title={Chronos: Learning the language of time series},
  author={Ansari, Abdul Fatir and Stella, Lorenzo and Turkmen, Caner and Zhang, Xiyuan and Mercado, Pedro and Shen, Huibin and Shchur, Oleksandr and Rangapuram, Syama Sundar and Arango, Sebastian Pineda and Kapoor, Shubham and others},
  journal={arXiv preprint arXiv:2403.07815},
  year={2024}
}

@article{bouten2014network,
  title={In-network quality optimization for adaptive video streaming services},
  author={Bouten, Niels and Latr{\'e}, Steven and Famaey, Jeroen and Van Leekwijck, Werner and De Turck, Filip},
  journal={IEEE Transactions on Multimedia},
  volume={16},
  number={8},
  pages={2281--2293},
  year={2014},
  publisher={IEEE}
}

@article{lim2021temporal,
  title={Temporal fusion transformers for interpretable multi-horizon time series forecasting},
  author={Lim, Bryan and Ar{\i}k, Sercan {\"O} and Loeff, Nicolas and Pfister, Tomas},
  journal={International Journal of Forecasting},
  volume={37},
  number={4},
  pages={1748--1764},
  year={2021},
  publisher={Elsevier}
}

@inproceedings{singh2022traffic,
  title={Traffic engineering: from isp to cloud wide area networks},
  author={Singh, Rachee and Bj{\o}rner, Nikolaj and Krishnaswamy, Umesh},
  booktitle={Proceedings of the Symposium on SDN Research},
  pages={50--58},
  year={2022}
}

@inproceedings{le2016joint,
  title={Joint capacity planning and operational management for sustainable data centers and demand response},
  author={Le, Tan N and Liu, Zhenhua and Chen, Yuan and Bash, Cullen},
  booktitle={Proceedings of the Seventh International Conference on Future Energy Systems},
  pages={1--12},
  year={2016}
}

@article{liu2023leaf,
  title={LEAF: Navigating Concept Drift in Cellular Networks},
  author={Liu, Shinan and Bronzino, Francesco and Schmitt, Paul and Bhagoji, Arjun Nitin and Feamster, Nick and Crespo, Hector Garcia and Coyle, Timothy and Ward, Brian},
  journal={Proceedings of the ACM on Networking},
  volume={1},
  number={CoNEXT2},
  pages={1--24},
  year={2023},
  publisher={ACM New York, NY, USA}
}

@inproceedings{zhang2024caravan,
  title={Caravan: Practical Online Learning of $\{$In-Network$\}$$\{$ML$\}$ Models with Labeling Agents},
  author={Zhang, Qizheng and Imran, Ali and Bardhi, Enkeleda and Swamy, Tushar and Zhang, Nathan and Shahbaz, Muhammad and Olukotun, Kunle},
  booktitle={18th USENIX Symposium on Operating Systems Design and Implementation (OSDI 24)},
  pages={325--345},
  year={2024}
}

@inproceedings{hensman2015scalable,
  title={Scalable variational Gaussian process classification},
  author={Hensman, James and Matthews, Alexander and Ghahramani, Zoubin},
  booktitle={Artificial Intelligence and Statistics},
  pages={351--360},
  year={2015},
  organization={PMLR}
}

\end{document}